\documentclass[10pt,twocolumn,letterpaper]{article}

\usepackage[pagenumbers]{cvpr}

\usepackage{microtype}

\usepackage{tabularx}
\usepackage{multirow}
\usepackage{wrapfig}
\usepackage{colortbl}
\usepackage{algorithm}
\usepackage{algorithmic}
\usepackage{tcolorbox}
\usepackage{etoc}

\definecolor{promptaccent}{HTML}{36566F}
\newtcolorbox{promptbox}[2][]{
  title={#2},
  colback=promptaccent!3!white,
  colframe=promptaccent!35!white,
  colbacktitle=promptaccent!10!white,
  coltitle=promptaccent,
  fonttitle=\small\bfseries,
  fontupper=\small\ttfamily,
  before upper={\raggedright\setlength{\parskip}{0pt}},
  boxrule=0.5pt,
  titlerule=0pt,
  arc=2mm,
  boxsep=0pt,
  left=3mm,
  right=3mm,
  top=2.5mm,
  bottom=2.5mm,
  toptitle=2mm,
  bottomtitle=2mm,
  before skip=9pt,
  after skip=9pt,
  #1
}

\tcbuselibrary{breakable}
\definecolor{caseaccent}{HTML}{38685C}
\newtcolorbox{casebox}[1]{
  breakable,
  title={#1},
  colback=caseaccent!3!white,
  colframe=caseaccent!55!white,
  colbacktitle=caseaccent!12!white,
  coltitle=caseaccent,
  fonttitle=\small\bfseries,
  fontupper=\small,
  before upper={\raggedright\setlength{\parskip}{4pt}},
  boxrule=0.6pt,
  titlerule=0pt,
  arc=0mm,
  boxsep=0pt,
  left=3mm,
  right=3mm,
  top=2mm,
  bottom=2mm,
  toptitle=2mm,
  bottomtitle=2mm,
  before skip=9pt,
  after skip=9pt
}

\makeatletter
\patchcmd{\cvprsection}{\large\bf}{\large\bf\hyphenpenalty=10000\exhyphenpenalty=10000}{}{}
\patchcmd{\cvprsubsection}{\elvbf}{\elvbf\hyphenpenalty=10000\exhyphenpenalty=10000}{}{}
\patchcmd{\cvprsubsubsection}{\tenbf}{\tenbf\hyphenpenalty=10000\exhyphenpenalty=10000}{}{}
\makeatother

\definecolor{cvprblue}{rgb}{0.21,0.49,0.74}
\usepackage[pagebackref,breaklinks,colorlinks,allcolors=cvprblue]{hyperref}

\def\confName{CVPR}
\def\confYear{2027}

\title{Seer: Guiding Small Deep-Research Agents with Shared Decision Models}

\author{Yuhao Zhou$^{1}$, Yang Zhou$^{1}$, Yuanxi Li$^{2}$, Jindi Lv$^{1}$, Dan Si$^{3}$, Qing Ye$^{1}$, Jiancheng Lv$^{1}$\thanks{Corresponding author: Jiancheng Lv.}\\
$^{1}$Sichuan University \quad $^{2}$University of Illinois at Urbana-Champaign \quad $^{3}$Lenovo Group Limited}

\begin{document}
\raggedbottom

\etocdepthtag.toc{main}

\maketitle

\begin{abstract}
Deep-research agents repeatedly decide what to search, read, and verify, often using large cloud-hosted language models for action proposal.
However, privacy and local control motivate running agents on users' devices, where limited memory and compute favor smaller models and complicate on-device training.
Yet inference-time guidance faces three obstacles: exploring alternatives through real execution adds tool calls, prompted self-evaluation is bounded by the agent's judgment, and model-specific guidance risks repeated adaptation costs.
We introduce \textbf{Seer}, an inference-time framework providing local guidance that separates the agent's action proposal from learned consequence prediction and value estimation.
From a shared corpus of real research trajectories, we train two models: a tool-response simulator on observed call--response pairs and a value estimator on histories labeled with discounted terminal correctness.
The simulator and estimator are trained offline and reused across agents without further training.
Moreover, Seer executes only the selected action and adds only real tool responses to the committed history, avoiding tool calls on discarded candidates at additional inference cost.
Across multiple deep-research benchmarks, Seer substantially improves agent performance for different language models.
For instance, GAIA accuracy rises from 13.6\% to 41.7\% for Qwen3.5-4B and from 22.3\% to 50.5\% for Qwen3.5-9B.
Furthermore, our analyses suggest that broader candidate coverage can compensate for imperfect simulation, while prediction and scoring errors limit the benefits of wider imagined search.
These findings highlight the potential of reusable models for prediction and evaluation to strengthen small research agents through extra local computation without the real execution \mbox{of the discarded candidates.}

\end{abstract}

\section{Introduction}
\label{sec:introduction}

Deep-research agents use language models to decide what to search, read, and verify, integrating evidence across sources to answer complex questions~\citep{deepresearchsurvey}.
In practice, OpenAI Deep Research~\citep{openaideepresearch} and Gemini Deep Research~\citep{geminideepresearch} offer multi-step web investigation and source-linked reports, while open efforts such as DeepResearcher~\citep{deepresearcher} and Tongyi DeepResearch~\citep{tongyideepresearch} advance research on sustained information seeking.
Meanwhile, benchmarks from GAIA's real-world reasoning and tool-use questions~\citep{gaia} to BrowseComp's challenging web information-seeking tasks~\citep{browsecomp} evaluate agents' ability to gather and integrate evidence.

For users applying these capabilities to sensitive tasks, keeping an agent's reasoning and decision-making on-device offers greater local control, a motivation reflected in privacy-oriented personal assistants~\citep{applefoundations}.
However, deploying these agents on users' devices requires managing both local computational resources and external tool-use costs.
Memory and compute constraints favor smaller language models~\citep{slmagents}, while tool-call fees make repeated searches and page retrievals an additional deployment expense~\citep{bats}.

\begin{figure*}[!t]
\centering
\begin{subfigure}[b]{0.3696\textwidth}
\centering
\includegraphics[width=\linewidth]{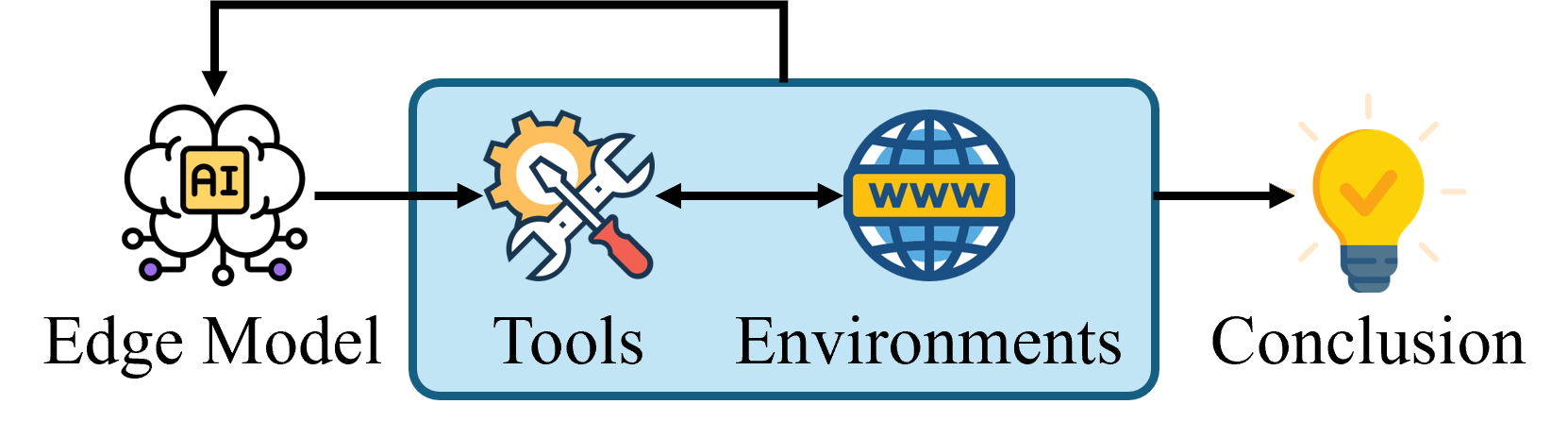}
\caption{High error rate.}
\label{fig:motivation-sub-1}
\end{subfigure}
\begin{subfigure}[b]{0.4144\textwidth}
\centering
\includegraphics[width=\linewidth]{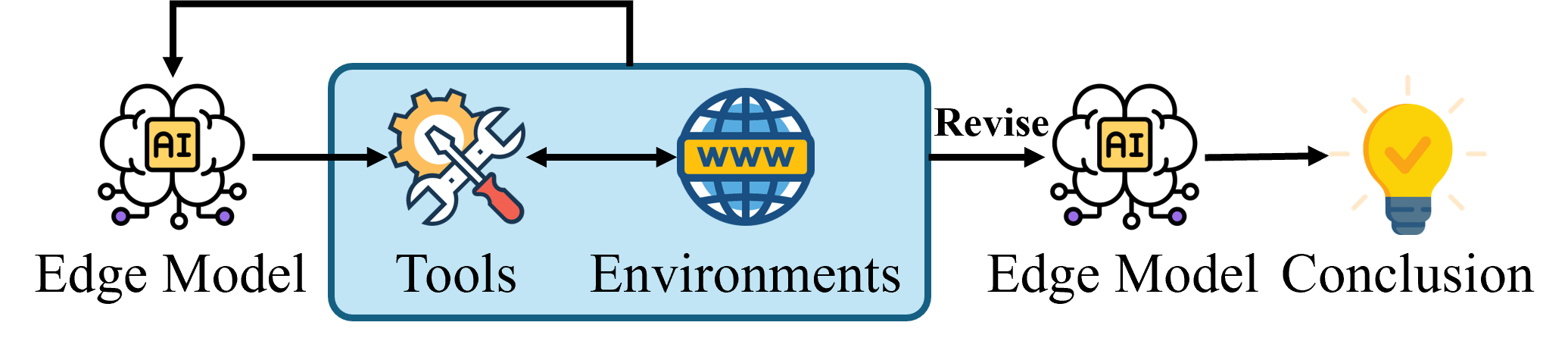}
\caption{Biased revision.}
\label{fig:motivation-sub-2}
\end{subfigure}
\begin{subfigure}[b]{0.3696\textwidth}
\centering
\vspace{5pt}
\includegraphics[width=\linewidth]{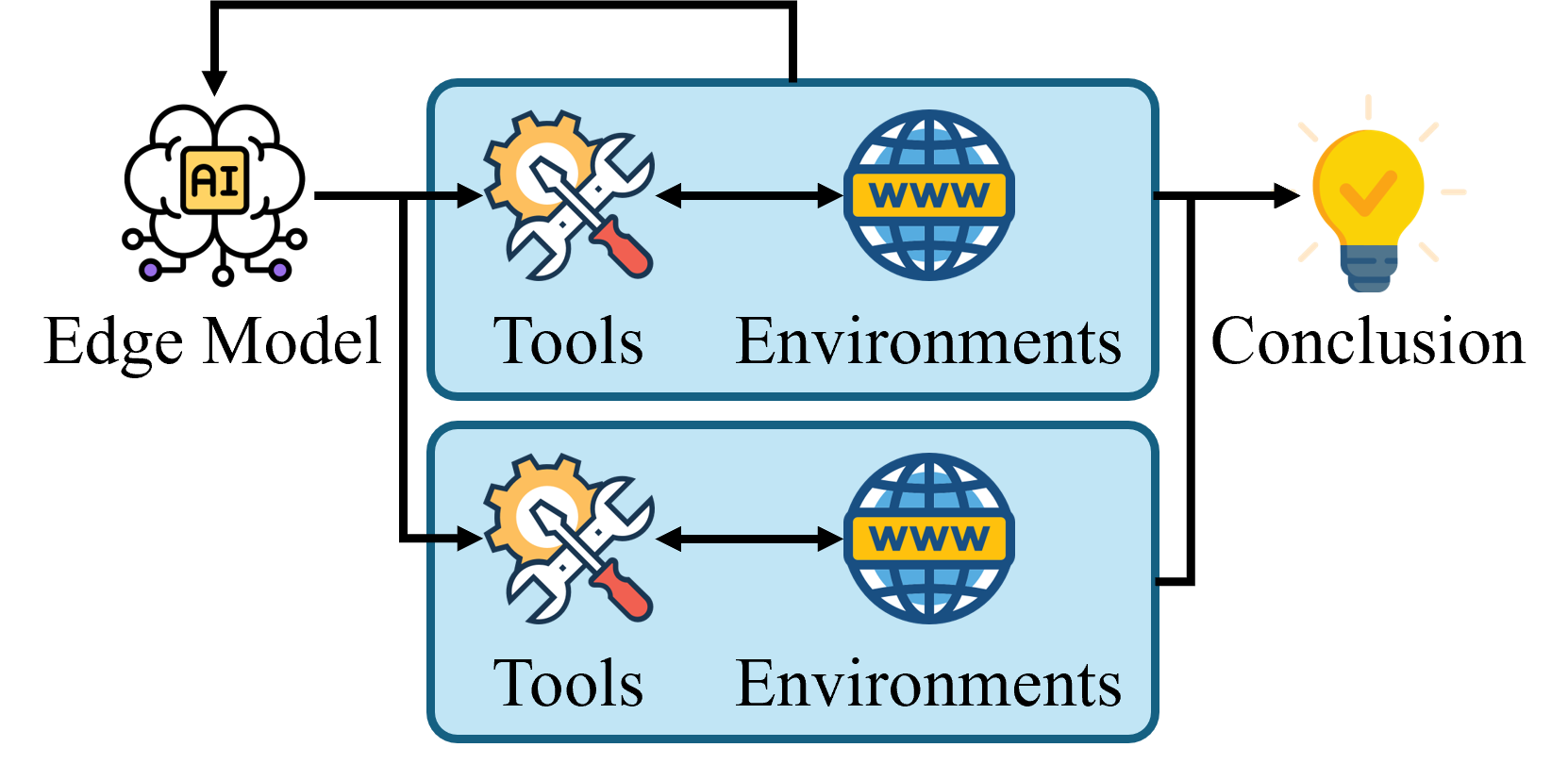}
\caption{Additional tool costs.}
\label{fig:motivation-sub-3}
\end{subfigure}
\begin{subfigure}[b]{0.4144\textwidth}
\centering
\vspace{5pt}
\includegraphics[width=\linewidth]{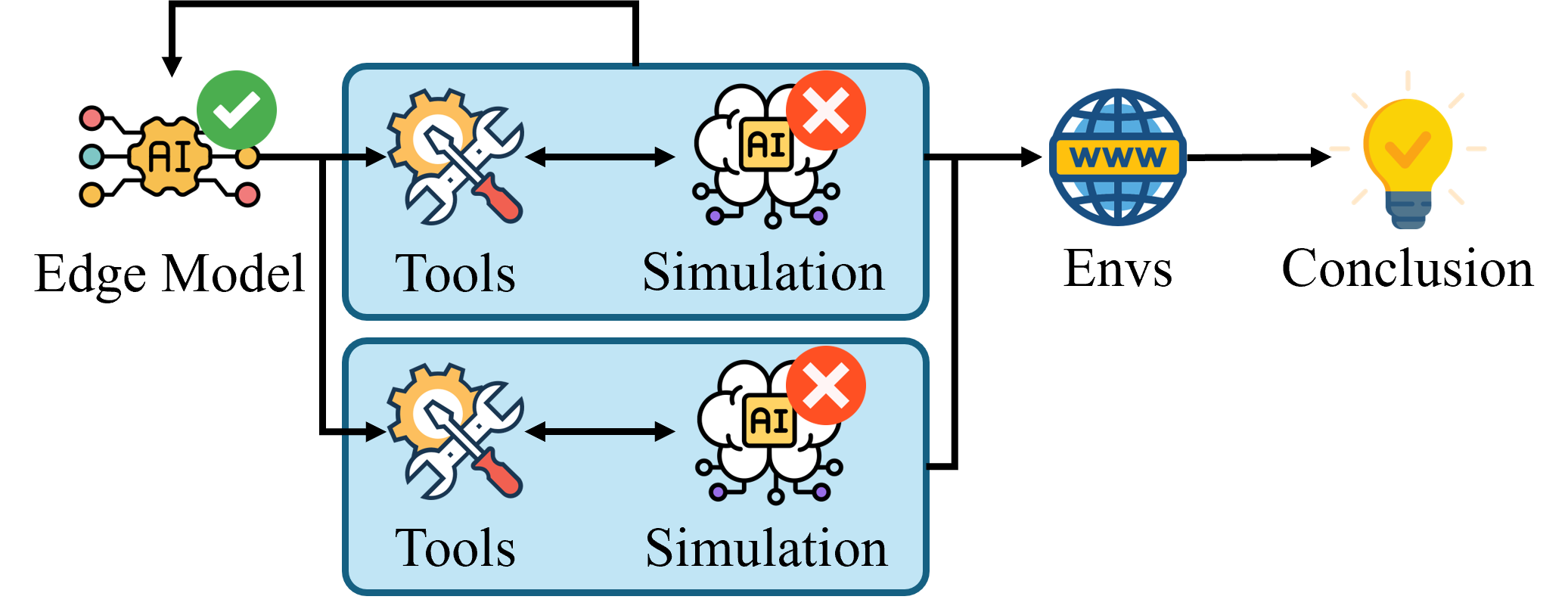}
\caption{Limited generalization across agent models.}
\label{fig:motivation-sub-4}
\end{subfigure}
\caption{How can we provide \textit{reusable} guidance that \textit{strengthens small agents} without model-specific retraining or the \textit{real execution} of every discarded candidate?}
\label{fig:motivation}
\end{figure*}

Under these constraints, training small models specifically for research is a natural way to improve agent performance.
However, collecting training data and optimizing each model require additional resources, making repeated adaptation costly for resource-constrained deployments.
Instead, inference-time methods offer an alternative by improving agent decisions without updating the deployed model.
Specifically, \textbf{output-level refinement}~\citep{selfconsistency,av,deepverifier} selects or revises completed answers but cannot redirect actions already executed.
\textbf{Decision-level evaluation}~\citep{lats,treeweb,agentrm,raggym} guides intermediate choices through search or learned critics, but evaluating branches through real execution incurs tool costs even for discarded candidates.
\textbf{Model-based lookahead}~\citep{rap,webdreamer,wma} reduces this overhead by evaluating predicted consequences before execution.
To avoid shifting adaptation costs to supporting components, such guidance should transfer across agent models without retraining.
Yet approaches relying on the agent's model for prompted prediction and scoring remain bounded by its capabilities~\citep{ondevicecontext}.
Figure~\ref{fig:motivation} summarizes these limitations.
The challenge is therefore to \textit{provide reusable guidance that strengthens small agents without model-specific retraining or the real execution of discarded candidates}.

We introduce \textbf{Seer}, an inference-time framework providing local guidance in deep research.
Our key idea is to \emph{decouple action proposal from tool-response prediction and value estimation to enable reusable guidance across agents}.
From a shared corpus of real research trajectories collected across tasks and agents, we train two models: a \emph{tool-response simulator} on observed call--response pairs and a \emph{value estimator} on interaction histories labeled with discounted terminal correctness.
These complementary supervision signals teach the simulator to predict what an action may reveal and the estimator to assess how the resulting history relates to eventual task success.
The simulator and estimator are trained offline and shared across agents powered by different models.
\textit{Deployment requires neither updating the agent's model nor retraining the simulator or estimator.}
Specifically, at inference, the agent proposes candidate actions, the simulator predicts their tool responses, and the estimator scores the resulting imagined histories to guide action selection.
Seer executes only the selected action and adds real tool responses to the committed history.
Candidate evaluation therefore uses additional inference computation rather than real tool calls on discarded alternatives, while the agent retains real rather than simulated responses as evidence.
Our contributions are threefold.

\begin{enumerate}[leftmargin=1.2em]
\item We propose Seer, an inference-time framework that decouples action proposal from shared tool-response prediction and history-level value estimation across agents. This design enables reusable guidance for frozen agents without any model-specific retraining or the real execution of the discarded candidates.

\item We develop a shared training pipeline that collects real trajectories across agents and tasks through Monte Carlo branching on existing datasets. The resulting large-scale training corpus provides complementary supervision: observed call--response pairs for the simulator and history prefixes labeled with discounted terminal correctness for the value estimator.

\item Experiments with different models across extensive benchmarks demonstrate substantial gains for small agents, with benefits extending to larger-model settings. Ablations and analyses reveal the trade-off between candidate coverage and imperfect prediction and scoring.
\end{enumerate}

\section{Related Work}
\label{sec:related}

\textbf{Research agents and tool use.}
Research agents couple a language model's parametric knowledge with information acquired through tools: WebGPT~\citep{webgpt} combines browser-assisted question answering with human feedback, ReAct~\citep{react} interleaves reasoning with actions, and Toolformer~\citep{toolformer} learns when to invoke APIs from self-supervised examples.
Within this setting, the task determines which observations matter: an agent must judge which query or goal-conditioned page visit is likely to reveal useful evidence, which creates two deployment costs—the local compute to run the model and the external tool calls to gather evidence.
Quantization~\citep{awq}, context management~\citep{ondevicecontext}, and tool budgeting~\citep{bats} address deployment costs, while compact research agents~\citep{agentcpmexplore,agentcpmreport,drvenus} target research capability.
Distillation~\citep{minillm,gkd} and reinforcement learning~\citep{searchr1,deepresearcher} strengthen the agent's model but require optimization for each adapted model.
Seer instead trains reusable decision models, complementing these efficiency-oriented approaches without updating or retraining the deployed agent.

\textbf{Inference-time guidance.}
Guidance can act on completed outputs or intervene at intermediate decisions, and the distinction matters because producing a candidate answer may already consume substantial tool calls.
Output-level methods select or revise solutions: self-consistency~\citep{selfconsistency} aggregates independently sampled reasoning paths, Self-Refine~\citep{selfrefine} alternates generation with feedback, and learned verifiers~\citep{verifiers} score candidates, with asymmetric verification~\citep{av} allocating a modest verification budget to select among expensive search trajectories.
These methods cannot recover tool calls already spent on discarded trajectories, as our BoN@4 and AV@4 baselines illustrate.
Decision-level methods intervene earlier: Tree of Thoughts~\citep{tot} explores intermediate reasoning states with self-evaluation, Reflexion~\citep{reflexion} uses verbal feedback from earlier attempts, LATS~\citep{lats} and web-agent tree search~\citep{treeweb} combine search with environmental feedback, and AgentRM~\citep{agentrm} learns a transferable reward model.
Process-supervision evaluators~\citep{stepverify,omegaprm,agentprm,raggym} further supply learned ranking signals, but branches explored through real interaction still incur tool calls even when discarded, and prompted self-evaluation depends on the agent's judgment.

\textbf{Model-based lookahead.}
A learned or prompted world model can instead predict aspects of a candidate's outcome before execution.
Dyna~\citep{dyna} integrates real experience with model-based planning, MuZero~\citep{muzero} learns latent dynamics together with planning-relevant predictions, and Dreamer-style agents~\citep{dreamerv3,mbpo} train behavior from imagined rollouts; for language agents, DreamGym~\citep{dreamgym} and DynaWeb~\citep{dynaweb} use predicted experience to train the agent itself.
At inference time, RAP~\citep{rap} repurposes an LLM as a world model within tree search, while WMA~\citep{wma} and WebDreamer~\citep{webdreamer} learn transition predictions for web navigation; SiRA~\citep{sira} plans with natural-language belief states, and MCP-Cosmos~\citep{mcpcosmos} couples world models with tool interfaces.
However, predicted experience in these works either trains the agent or guides a single navigation decision, and avoiding real candidate execution does not by itself ensure that both prediction and evaluation transfer across agent models without retraining.
To enable this reuse, Seer separates call-level response prediction and history-level evaluation from the agent model, training the evaluator on discounted terminal correctness and reusing both checkpoints across agents.
It executes only the selected action and retains real responses as evidence, trading additional local inference for candidate comparison without tool calls on discarded candidates.

\section{Problem Formulation}
\label{sec:formulation}

Given a request $q$, a deep-research agent $\mathcal{G}$ uses a language model to decide what to search, read, and verify, integrating evidence across sources to produce an answer or report $\hat{a}$~\citep{deepresearchsurvey}.
During a run, the agent maintains a committed interaction history $s_t$, which initially contains $q$ and the agent instructions.
At round $t$, it uses this history to select an action $a_t$ that either submits a final answer without tool calls or issues one or more tool calls.
To describe this choice mathematically, let $\pi_{\mathcal{G}}(\cdot\mid s_t)$ denote the agent's distribution over actions given the current history.

Let $\mathcal{E}$ denote the real tool-execution operator, which invokes the tools specified in $a_t$ with their arguments in the current external environment and returns the responses $o_t$.
Action selection and the history update for a nonterminal interaction are then written as
\begin{equation}
a_t \sim \pi_{\mathcal{G}}(\cdot\mid s_t), \quad
o_t = \mathcal{E}(a_t), \quad
s_{t+1}=s_t\oplus(a_t,o_t),
\end{equation}
where $\oplus$ denotes concatenation.
Thus, the next action is selected using a history that includes the previous history, the selected action, and its real tool responses.
A final-answer submission instead terminates the run with $o_t=\varnothing$.

We denote the full execution record by $\xi$ and assess its outcome using a task-dependent terminal quality score $R_{\mathrm{term}}(\xi;q)\in[0,1]$.
For the question-answering tasks considered here, this score is 1 for a correct final answer and 0 otherwise, including unanswered tasks.
Gold answers are used only for evaluation and offline supervision, not for inference-time decisions.
For requests $q\sim\mathcal{D}$ drawn from the task distribution and traces $\xi$ generated by $\mathcal{G}$ through real tool interactions, the objective is
\begin{equation}
\label{eq:research-objective}
\max_{\mathcal{G}}\;
\mathbb{E}_{q\sim\mathcal{D},\xi}\big[R_{\mathrm{term}}(\xi;q)\big].
\end{equation}

When the agent's reasoning and decision-making run on-device, this objective is constrained by local model memory and the cost of real tool calls to external services.
We denote the total number of locally deployed model parameters by $c_{\mathrm{param}}$ and the number of real tool calls by $c_{\mathrm{tool}}$, and collect them in
\begin{equation}
\mathbf{c}(\mathcal{G},\xi)=
\big(c_{\mathrm{param}}(\mathcal{G}),\,
     c_{\mathrm{tool}}(\xi)\big)^{\top}.
\end{equation}
Given the corresponding budget vector $\mathbf{B}$, the on-device problem becomes
\begin{equation}
\label{eq:ondevice-objective}
\max_{\mathcal{G}}\;
\mathbb{E}_{q\sim\mathcal{D},\xi}\big[R_{\mathrm{term}}(\xi;q)\big]
\quad\text{s.t.}\quad
\mathbf{c}(\mathcal{G},\xi)\preceq\mathbf{B},
\end{equation}
where $\preceq$ denotes componentwise inequality.
The per-run count $c_{\mathrm{tool}}$ includes all real calls recorded in $\xi$, including those made for discarded candidates, with batched calls counted individually.
A run stops on submission or when the budget is exhausted.

To compare candidates without increasing $c_{\mathrm{tool}}$, let $\widehat{\mathcal{E}}$ denote a local approximation to tool responses, following planning with predicted action outcomes~\citep{rap,webdreamer}.
For a candidate action $a^{(i)}$ at history $s_t$, the predicted response and imagined history are
\begin{equation}
\label{eq:imagined-history}
\hat{o}^{(i)}=\widehat{\mathcal{E}}(a^{(i)}),\qquad
\hat{s}^{(i)}=s_t\oplus(a^{(i)},\hat{o}^{(i)}),
\end{equation}
with $\hat{o}^{(i)}=\varnothing$ for final-answer candidates.
These imagined histories support candidate comparison, while the committed update still uses $o_t=\mathcal{E}(a_t)$.
Our focus is therefore to improve the objective in Eq.~(\ref{eq:ondevice-objective}) for a fixed agent model using predicted outcomes for action selection, while satisfying the same resource budgets.
Appendix~\ref{app:notation} summarizes the notation used throughout the paper.

\begin{figure*}[t]
\centering
\includegraphics[width=0.85\textwidth]{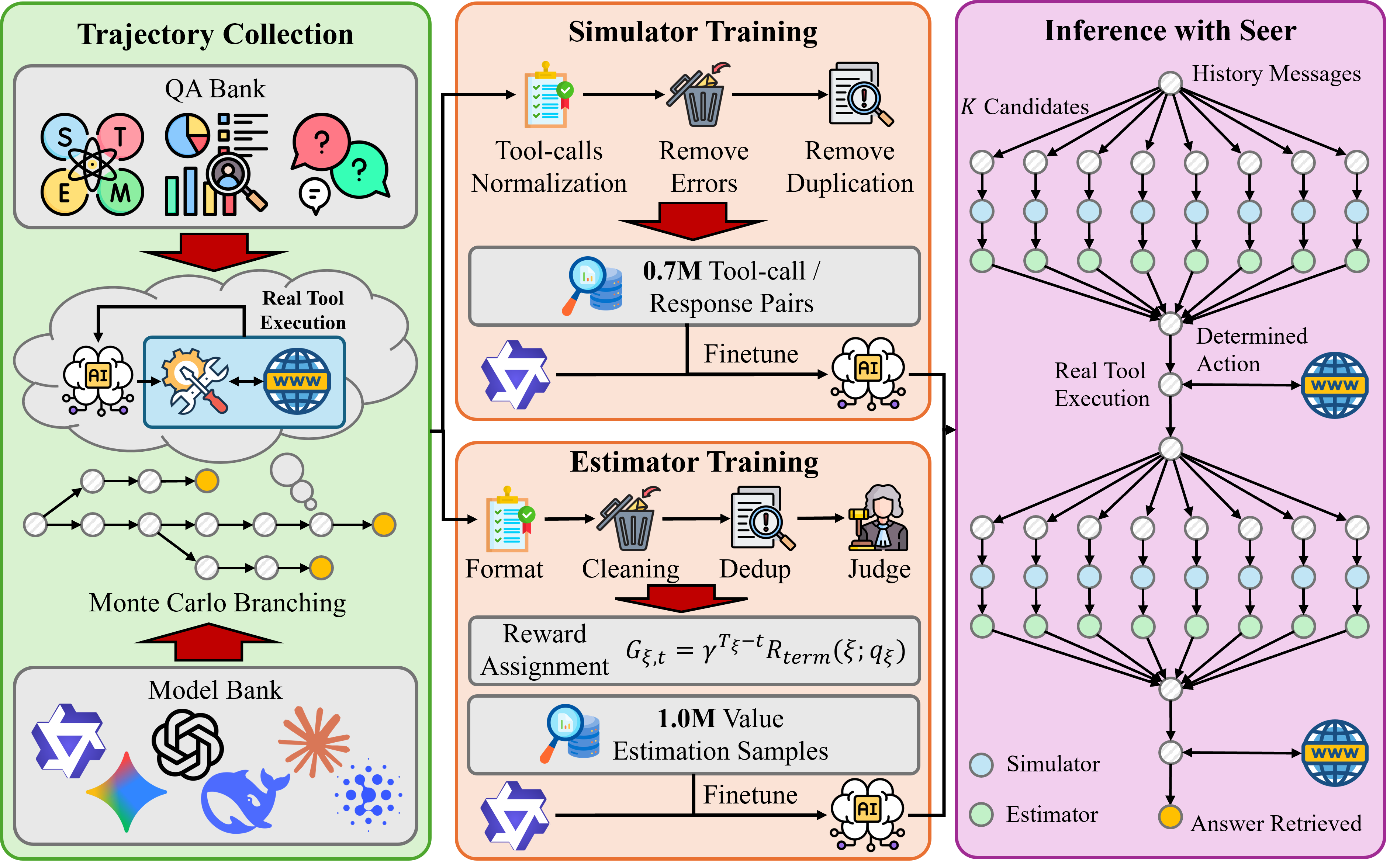}
\caption{Seer overview. Monte Carlo branching collects real trajectories to train the simulator on tool responses and the estimator on discounted terminal correctness. At inference, these shared models guide search over imagined histories, while only selected actions are executed to obtain real evidence.}
\label{fig:overview}
\end{figure*}

\section{Our Approach}
\label{sec:approach}
\suppressfloats[t]

The response approximation in \S\ref{sec:formulation} leaves two questions: what might a candidate action reveal, and how useful is the resulting history for completing the task?
Seer addresses them with a \emph{tool-response simulator} $W_\phi$ and a \emph{value estimator} $V_\theta$ supervised by observed responses and discounted terminal correctness, respectively, while the agent's frozen model proposes actions (Figure~\ref{fig:overview}).
We refer to these two components jointly as the \emph{decision models}, and individually as the simulator and estimator.

\subsection{Decoupling Prediction from Evaluation}
\label{sec:decision-models}

\textbf{Predicting tool responses.}
Let $\pi$ denote the frozen agent model's action-proposal distribution and $\operatorname{calls}(a)$ the tool invocations in action $a$.
For each call $c$, containing a tool name and arguments, $W_\phi$ generates a response $\hat{o}_c$ conditioned only on that call and a tool-specific instruction.
Collecting these responses in call order implements the local approximation
\begin{equation}
\label{eq:seer-simulation}
\hat{o}_c\sim W_\phi(\cdot\mid c),\qquad
\widehat{\mathcal{E}}(a)=\bigl(\hat{o}_c\bigr)_{c\in\operatorname{calls}(a)}.
\end{equation}
This interface omits conversation history and agent identity, allowing reuse across agents with a common tool schema.

\textbf{Evaluating candidate histories.}
Predicting a response does not establish its usefulness; $V_\theta$ therefore evaluates the resulting history using supervision from terminal task outcomes.
For each proposal $a^{(i)}\sim\pi(\cdot\mid s_t)$, we construct the imagined history as in Eq.~(\ref{eq:imagined-history}) and evaluate it with $V_\theta$:
\begin{equation}
\label{eq:seer-evaluation}
\hat{s}^{(i)}=s_t\oplus\bigl(a^{(i)},\widehat{\mathcal{E}}(a^{(i)})\bigr),\qquad
v^{(i)}=V_\theta(\hat{s}^{(i)}).
\end{equation}
For final-answer candidates, $\widehat{\mathcal{E}}(a^{(i)})=\varnothing$, so the estimator scores the history with the proposed answer appended.
Search uses these scores to change the overall action-selection distribution $\pi_{\mathcal{G}}$, not the proposal distribution $\pi$.
Both learned components are trained offline and reused across agents without further training.

\subsection{Learning from Real Research Trajectories}
\label{sec:training}

We collect a shared corpus using agents powered by different frozen language models on training questions disjoint from the evaluation questions.
Inspired by Monte Carlo search for process-supervision collection~\citep{omegaprm}, we branch from intermediate histories to collect alternative continuations, all through real tool execution.
The same trajectories thus supply two complementary signals: observed responses for $W_\phi$ and discounted terminal correctness for $V_\theta$.

\textbf{Learning Tool-Response Prediction}:
\label{sec:world}
We extract individual call--response pairs from the corpus, normalize calls, remove infrastructure failures, and deduplicate to obtain $\mathcal{D}_W=\{(c_i,o_i)\}_{i=1}^{N_W}$ with $N_W$ retained pairs.
Unlike value supervision, response supervision does not depend on terminal correctness: even an unsuccessful trajectory can contain valid tool outputs.

Starting from a pretrained autoregressive language model, we optimize its trainable parameters $\phi$ using a response-only negative log-likelihood:
\begin{equation}
\mathcal{L}_W(\phi)=
-\mathbb{E}_{(c,o)\sim\mathcal{D}_W}
\sum_{\ell=1}^{|o|}\log W_\phi(o_\ell\mid c,o_{<\ell}),
\end{equation}
where $o_\ell$ is response token $\ell$ and $o_{<\ell}$ denotes its preceding tokens.
The tool-specific instruction and call are omitted from the loss; training and inference use the same input template (Appendix~\ref{app:world}).

\textbf{Learning Outcome-Supervised Value Estimation}:
\label{sec:value}
For value supervision, we deduplicate trajectories with submitted answers, excluding unanswered error or timeout runs.
A judge compares each submission with its gold answer to assign $R_{\mathrm{term}}(\xi;q_\xi)\in\{0,1\}$; gold answers enter only label construction, never the estimator input.
For a trajectory $\xi$ with question $q_\xi$ and terminal history index $T_\xi$, we label each retained prefix by discounted terminal correctness:
\begin{equation}
\mathcal{D}_V=\{(s_{\xi,t},G_{\xi,t})\}_{\xi,t},\qquad
G_{\xi,t}=\gamma^{T_\xi-t}R_{\mathrm{term}}(\xi;q_\xi),
\end{equation}
where $0<\gamma<1$ discounts remaining interaction steps, not tokens or individual tool calls.
Incorrect trajectories receive zero throughout; correct ones receive increasing targets toward completion.
Consequently, $V_\theta$ ranks histories by outcome and proximity to completion rather than estimating a calibrated success probability for a deployment model.

We replace a pretrained language model's head with an HL-Gauss classification head~\citep{hlgauss}, with $\theta$ denoting the resulting model's trainable parameters.
Let $\beta_0<\cdots<\beta_K$ delimit $K$ uniform bins over an interval covering the targets, with centers $z_b=(\beta_{b-1}+\beta_b)/2$.
For a target $G$, Gaussian smoothing with standard deviation $\sigma$ assigns the normalized bin masses
\begin{equation}
\label{eq:seer-hlgauss}
y_b(G)=
\frac{\Phi\!\left((\beta_b-G)/\sigma\right)-\Phi\!\left((\beta_{b-1}-G)/\sigma\right)}
{\Phi\!\left((\beta_K-G)/\sigma\right)-\Phi\!\left((\beta_0-G)/\sigma\right)},
\end{equation}
where $\Phi$ is the standard normal cumulative distribution function.
Given predicted bin probabilities $p_\theta(b\mid s)$, we minimize soft-label cross-entropy and decode by expectation:
\begin{equation}
\begin{aligned}
\mathcal{L}_V(\theta)&=
-\mathbb{E}_{(s,G)\sim\mathcal{D}_V}\sum_{b=1}^{K}
y_b(G)\log p_\theta(b\mid s),\\
V_\theta(s)&=\sum_{b=1}^{K}z_b p_\theta(b\mid s).
\end{aligned}
\end{equation}
For efficient training, each trajectory is encoded once in a causal forward pass, with scores read at its labeled prefix endpoints.
This requires token-identical shared prefixes and exclusion of trajectories beyond the context limit.
Training and validation are split by question, keeping all trajectories and prefixes for the same question together.

\subsection{Imagined Search and Real Execution}
\label{sec:planner}

At each real round, Seer builds a fresh tree rooted at the committed history $s_t$ and runs $N_{\mathrm{sim}}$ simulations using predictor upper confidence bound applied to trees (PUCT)~\citep{rosin2011,alphazero}.
Algorithm~\ref{alg:seer-search} in Appendix~\ref{app:planner} gives the complete round, with candidate width $k$ and depth cap $d_{\max}$.
Here, a search simulation is one tree traversal and value backup, distinct from predicting an individual tool response with the simulator.

\textbf{Candidate expansion.}
At an expandable history $\bar{s}$, we independently sample $m$ valid actions from $\pi(\cdot\mid\bar{s})$, with $1\le m\le k$, and batch their tool calls through $W_\phi$.
The resulting imagined histories become children with uniform edge priors $P(e)=1/m$; final-answer children are terminal.
Because $V_\theta$ is trained on real histories, simulation errors can distort its rankings, motivating the depth cap.

\textbf{Selection and backup.}
For an edge $e$ with parent node $u$, selection uses
\begin{equation}
\label{eq:seer-puct}
\operatorname{score}(e)=Q(e)+c_{\mathrm{puct}}P(e)
\frac{\sqrt{N(u)+1}}{1+N(e)},
\end{equation}
where $N(u)$ and $N(e)$ are visit counts, $Q(e)$ is the mean backed-up value, and $c_{\mathrm{puct}}$ controls exploration.
Each simulation scores its endpoint with $V_\theta$ and backs up that value along the traversed path, including when the endpoint is terminal or depth-capped.
Thus, terminal candidates are evaluated by the learned estimator, not by real execution or gold-answer judgments.

\textbf{Real execution.}
After search, we commit the action on the most-visited root edge:
\begin{equation}
\label{eq:seer-commit}
e^*\in\arg\max_{e\in\mathrm{root\ edges}}N(e),\qquad
a_t=\operatorname{action}(e^*).
\end{equation}
A final-answer submission terminates the run; otherwise, we execute $a_t$ and update $s_{t+1}=s_t\oplus(a_t,\mathcal{E}(a_t))$.
Imagined branches are discarded, so only real responses enter the next round's root history.
Accordingly, Seer's real tool-call count is given by
\begin{equation}
c_{\mathrm{tool}}(\xi)=\sum_t\bigl|\operatorname{calls}(a_t)\bigr|,
\end{equation}
where final-answer actions contribute zero.
Discarded candidates incur local inference but no real calls.
Total calls can nevertheless differ from the unassisted agent because action choices and run length change.
In addition, Seer uses additional local computation and model memory for candidate comparison without relaxing the resource budgets.

\section{Experiments}
\label{sec:experiments}

\begin{table}[t]
\centering
\small
\caption{Preliminary comparison of RAP@4 and Seer ($k=4$) using average accuracy across benchmarks (\%). RAP@4 is omitted from later tables as it barely improves on the Base Agent.}
\label{tab:rap}
\vspace{-5pt}
\setlength{\tabcolsep}{6pt}
\begin{tabular*}{\linewidth}{@{\extracolsep{\fill}}lrr>{\columncolor{black!7}}r@{}}
\toprule
Model & Base & RAP@4 & Seer \\
\midrule
Qwen3.5-4B & \underline{8.3} & 8.0 & \textbf{18.8} \\
Qwen3.5-9B & \underline{14.2} & 13.9 & \textbf{25.2} \\
\bottomrule
\end{tabular*}
\vspace{-10pt}
\end{table}

\subsection{Experimental Setup}
\begin{table*}[t]
\centering
\small
\caption{Comparisons of final-answer accuracy (\%) across Qwen and Gemma agents using Base Agent, BoN@4, AV@4, and Seer ($S_k$, candidate width $k$). Seer improves over Base Agent in 26 of 30 settings at $k=32$, reusing the simulator and estimator across agents.}
\label{tab:main}
\vspace{-5pt}
\renewcommand{\arraystretch}{1.15}
\setlength{\tabcolsep}{4pt}
\begin{tabular*}{0.86\linewidth}{@{\extracolsep{\fill}}l*{3}{c}*{2}{>{\columncolor{black!7}}c}@{\hspace{12pt}}*{3}{c}*{2}{>{\columncolor{black!7}}c}@{\hspace{12pt}}*{3}{c}*{2}{>{\columncolor{black!7}}c}@{}}
\toprule
 & \multicolumn{5}{c}{Qwen3.5-4B} & \multicolumn{5}{c}{Qwen3.5-9B} & \multicolumn{5}{c}{Qwen3.5-35B-A3B} \\
\cmidrule(lr){2-6}\cmidrule(lr){7-11}\cmidrule(lr){12-16}
Benchmark & Base & BoN & AV & $S_4$ & $S_{32}$ & Base & BoN & AV & $S_4$ & $S_{32}$ & Base & BoN & AV & $S_4$ & $S_{32}$ \\
\midrule
GAIA & 13.6 & 21.4 & 26.2 & \underline{29.1} & \textbf{41.7} & 22.3 & \underline{35.9} & 35.0 & \underline{35.9} & \textbf{50.5} & \underline{57.3} & 53.4 & 47.6 & \textbf{62.1} & \underline{57.3} \\
BC-200 & 1.0 & 4.5 & 3.5 & \underline{5.0} & \textbf{7.0} & 2.5 & 5.5 & \underline{6.5} & 5.5 & \textbf{9.5} & 22.0 & \underline{24.0} & \underline{24.0} & \textbf{24.5} & 22.0 \\
SEAL-0 & 4.5 & 9.9 & 8.1 & \underline{10.8} & \textbf{16.2} & 10.8 & \underline{23.4} & 18.9 & 22.5 & \textbf{26.1} & 27.0 & 28.8 & \underline{33.3} & 31.5 & \textbf{34.2} \\
XBench-DS & 20.0 & 36.0 & 35.0 & \underline{41.0} & \textbf{43.0} & 28.0 & 38.0 & 38.0 & \underline{47.0} & \textbf{60.0} & 64.0 & \underline{67.0} & 64.0 & 66.0 & \textbf{69.0} \\
BC-ZH & 2.4 & 7.6 & 5.2 & \underline{8.0} & \textbf{9.7} & 7.6 & 11.4 & 13.5 & \underline{15.2} & \textbf{16.3} & 36.3 & \underline{37.7} & 34.9 & 36.0 & \textbf{43.3} \\
\bottomrule
\end{tabular*}

\smallskip
\begin{tabular*}{0.86\linewidth}{@{\extracolsep{\fill}}l*{3}{c}*{2}{>{\columncolor{black!7}}c}@{\hspace{12pt}}*{3}{c}*{2}{>{\columncolor{black!7}}c}@{\hspace{12pt}}*{3}{c}*{2}{>{\columncolor{black!7}}c}@{}}
 & \multicolumn{5}{c}{Qwen3.6-35B-A3B} & \multicolumn{5}{c}{Gemma-4-e2b-it} & \multicolumn{5}{c}{Gemma-4-e4b-it} \\
\cmidrule(lr){2-6}\cmidrule(lr){7-11}\cmidrule(lr){12-16}
Benchmark & Base & BoN & AV & $S_4$ & $S_{32}$ & Base & BoN & AV & $S_4$ & $S_{32}$ & Base & BoN & AV & $S_4$ & $S_{32}$ \\
\midrule
GAIA & 64.1 & 75.7 & 73.8 & \underline{78.6} & \textbf{79.6} & 20.4 & 23.3 & 25.2 & \underline{30.1} & \textbf{31.1} & 21.4 & 25.2 & \underline{30.1} & 24.3 & \textbf{36.9} \\
BC-200 & 20.5 & 35.0 & 35.0 & \underline{36.0} & \textbf{37.5} & \underline{1.5} & 1.0 & 1.0 & \textbf{2.5} & 1.0 & 0.5 & \textbf{1.5} & \textbf{1.5} & \underline{1.0} & \textbf{1.5} \\
SEAL-0 & \textbf{29.7} & 20.7 & \underline{28.8} & 24.3 & 27.0 & 1.8 & \underline{5.4} & 4.5 & 4.5 & \textbf{6.3} & 6.3 & \textbf{9.0} & \underline{8.1} & 5.4 & 7.2 \\
XBench-DS & 61.0 & \textbf{76.0} & 70.0 & 73.0 & \underline{74.0} & 21.0 & 21.0 & \underline{29.0} & 28.0 & \textbf{31.0} & 23.0 & 28.0 & \underline{39.0} & 27.0 & \textbf{40.0} \\
BC-ZH & 27.0 & 35.3 & \underline{36.0} & 35.3 & \textbf{38.8} & 2.1 & 1.0 & \underline{2.8} & 1.7 & \textbf{3.8} & 1.4 & 1.0 & \textbf{4.2} & \underline{3.8} & 3.5 \\
\bottomrule
\end{tabular*}
\vspace{-5pt}
\end{table*}

\begin{table*}[t]
\centering
\small
\caption{Comparisons of final-answer accuracy (\%) on larger agents. Seer transfers to 122B and 397B agents without retraining, with $S_{32}$ outperforming Base Agent in all cases.}
\label{tab:large-models}
\vspace{-5pt}
\renewcommand{\arraystretch}{1.15}
\setlength{\tabcolsep}{5pt}
\begin{tabular*}{0.86\linewidth}{@{\extracolsep{\fill}}l*{3}{c}*{2}{>{\columncolor{black!7}}c}@{\hspace{10pt}}*{3}{c}*{2}{>{\columncolor{black!7}}c}@{}}
\toprule
 & \multicolumn{5}{c}{Qwen3.5-122B-A10B} & \multicolumn{5}{c}{Qwen3.5-397B-A17B} \\
\cmidrule(lr){2-6}\cmidrule(lr){7-11}
Benchmark & Base & BoN & AV & $S_4$ & $S_{32}$ & Base & BoN & AV & $S_4$ & $S_{32}$ \\
\midrule
GAIA & 67.0 & 64.1 & 61.2 & \underline{68.0} & \textbf{71.8} & 68.9 & \underline{74.8} & 69.9 & \textbf{76.7} & 72.8 \\
BC-200 & 31.5 & 32.5 & \underline{35.0} & \underline{35.0} & \textbf{39.0} & 34.5 & \underline{37.5} & \underline{37.5} & 37.0 & \textbf{42.5} \\
SEAL-0 & 32.4 & 28.8 & 31.5 & \underline{34.2} & \textbf{36.9} & 31.5 & 36.9 & 36.0 & \underline{37.8} & \textbf{40.5} \\
XBench-DS & 66.0 & 70.0 & 61.0 & \underline{71.0} & \textbf{72.0} & 75.0 & 73.0 & 75.0 & \underline{79.0} & \textbf{80.0} \\
BC-ZH & 39.4 & 38.1 & 36.3 & \underline{40.5} & \textbf{43.6} & 50.2 & 49.8 & 49.8 & \textbf{55.7} & \underline{55.4} \\
\bottomrule
\end{tabular*}
\end{table*}

\begin{table*}[t]
\centering
\small
\caption{Comparisons of cumulative component additions on GAIA across agents (accuracy \%, $k=4$). Adding the estimator improves over prompted scoring, while subsequently adding the simulator avoids real calls for discarded candidates at an accuracy cost.}
\label{tab:ablation}
\vspace{-5pt}
\renewcommand{\arraystretch}{1.2}
\setlength{\tabcolsep}{5pt}
\begin{tabularx}{\linewidth}{@{}>{\raggedright\arraybackslash}p{0.22\linewidth}*{2}{>{\hsize=0.65\hsize\linewidth=\hsize\centering\arraybackslash}X}*{4}{>{\hsize=1.35\hsize\linewidth=\hsize\centering\arraybackslash}X}*{2}{>{\hsize=0.65\hsize\linewidth=\hsize\centering\arraybackslash}X}@{}}
\toprule
 & \multicolumn{5}{c}{Qwen3.5} & Qwen3.6 & \multicolumn{2}{c}{Gemma-4} \\
\cmidrule(lr){2-6}\cmidrule(lr){7-7}\cmidrule(lr){8-9}
Variant & 4B & 9B & \mbox{35B-A3B} & \mbox{122B-A10B} & \mbox{397B-A17B} & \mbox{35B-A3B} & e2b-it & e4b-it \\
\midrule
Base + search ($k=4$) & 16.5 & \underline{44.7} & \underline{62.1} & \underline{71.8} & \underline{73.8} & 74.8 & 26.2 & 19.4 \\
\quad + estimator & \textbf{38.8} & \textbf{45.6} & \textbf{66.0} & \textbf{72.8} & 71.8 & \underline{75.7} & \textbf{31.1} & \textbf{29.1} \\
\rowcolor{black!7}
\quad + simulator (Seer) & \underline{29.1} & 35.9 & \underline{62.1} & 68.0 & \textbf{76.7} & \textbf{78.6} & \underline{30.1} & \underline{24.3} \\
\quad \textit{Direct scoring} & 23.3 & 31.1 & 61.2 & 68.0 & 72.8 & 76.7 & 28.2 & 21.4 \\
\bottomrule
\end{tabularx}
\vspace{-10pt}
\end{table*}

\begin{figure*}[t]
\centering
\captionsetup{skip=3pt}
\captionsetup[subfigure]{font=scriptsize,skip=-2pt,belowskip=0pt}
\begin{subfigure}[b]{0.31\textwidth}
\centering
\includegraphics[width=\linewidth]{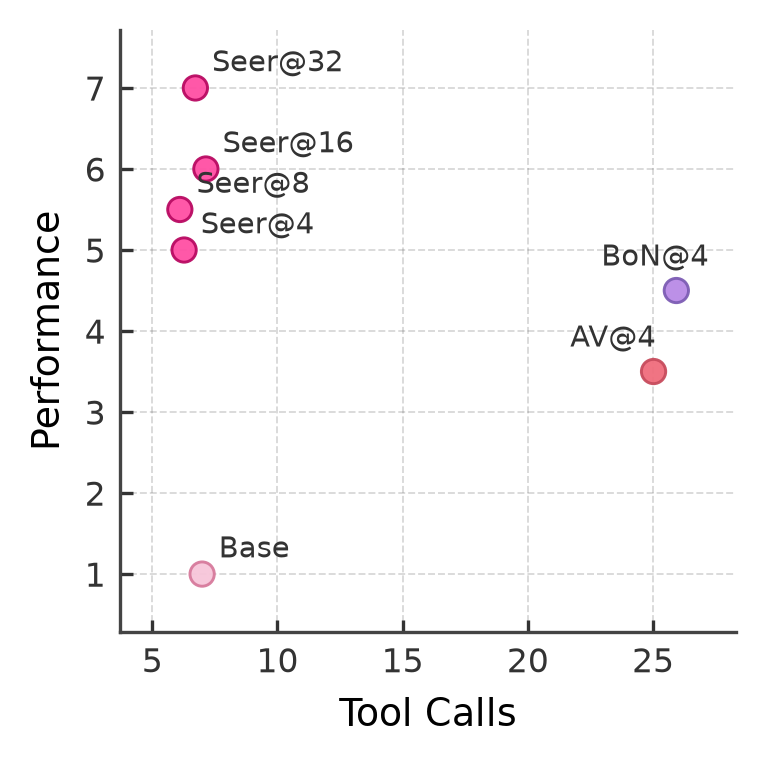}
\caption{Qwen3.5-4B}
\end{subfigure}\hfill
\begin{subfigure}[b]{0.31\textwidth}
\centering
\includegraphics[width=\linewidth]{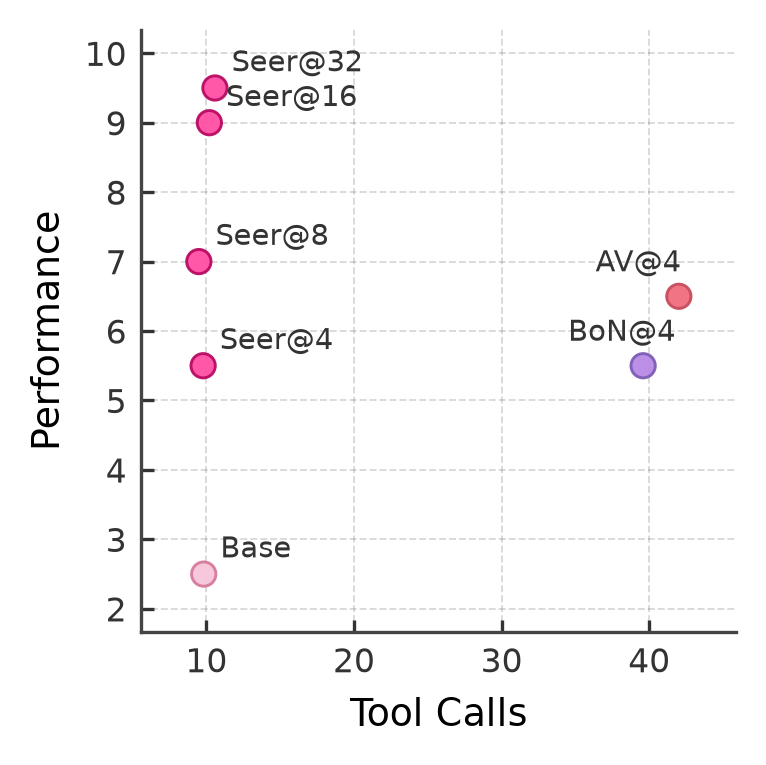}
\caption{Qwen3.5-9B}
\end{subfigure}\hfill
\begin{subfigure}[b]{0.31\textwidth}
\centering
\includegraphics[width=\linewidth]{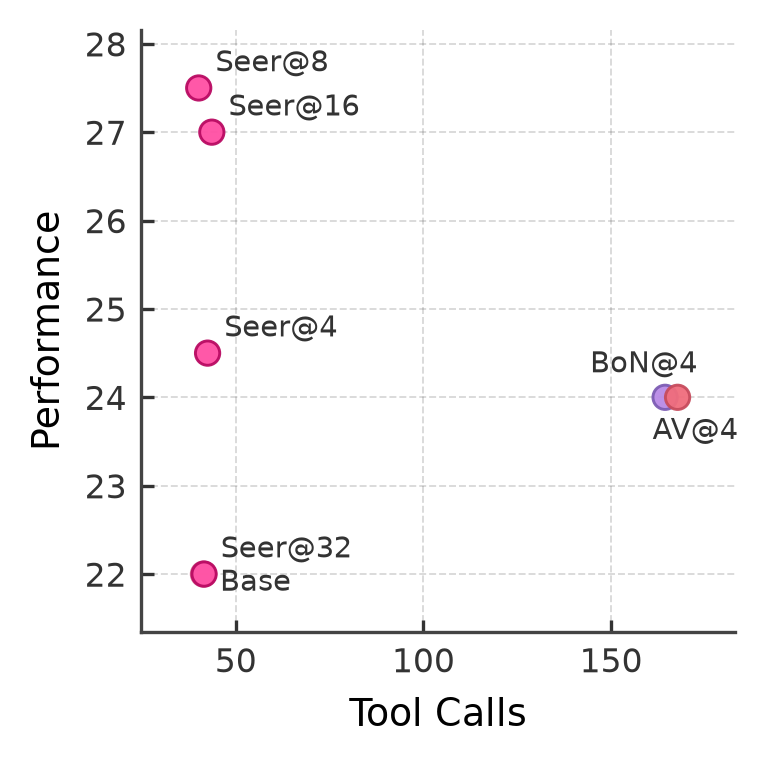}
\caption{Qwen3.5-35B-A3B}
\end{subfigure}
\caption{Comparisons of final-answer accuracy and real tool-call counts on BrowseComp-200 across agents. The vertical axis (Performance) reports accuracy (\%). Seer achieves the highest accuracy while keeping tool-call counts close to those of Base Agent.}
\label{fig:tool-cost}
\end{figure*}

\begin{table*}[t]
\centering
\small
\caption{Comparisons of Seer accuracy (\%) across candidate widths $k$ on all five benchmarks, with the best and second-best results within each model and benchmark shown in bold and underlined, respectively, including ties. Larger $k$ does not consistently improve performance, highlighting the need to select a width for each agent and task.}
\label{tab:ksweep}
\vspace{-5pt}
\renewcommand{\arraystretch}{1.2}
\setlength{\tabcolsep}{4pt}
\begin{tabular*}{\linewidth}{@{\extracolsep{\fill}}l*{4}{c}@{\hspace{8pt}}*{4}{c}@{\hspace{8pt}}*{4}{c}@{\hspace{8pt}}*{4}{c}@{}}
\toprule
 & \multicolumn{4}{c}{Qwen3.5-4B} & \multicolumn{4}{c}{Qwen3.5-9B} & \multicolumn{4}{c}{Qwen3.5-35B-A3B} & \multicolumn{4}{c}{Qwen3.5-122B-A10B} \\
\cmidrule(lr){2-5}\cmidrule(lr){6-9}\cmidrule(lr){10-13}\cmidrule(lr){14-17}
Benchmark & $k{=}4$ & $k{=}8$ & $k{=}16$ & $k{=}32$ & $k{=}4$ & $k{=}8$ & $k{=}16$ & $k{=}32$ & $k{=}4$ & $k{=}8$ & $k{=}16$ & $k{=}32$ & $k{=}4$ & $k{=}8$ & $k{=}16$ & $k{=}32$ \\
\midrule
GAIA & 29.1 & 35.9 & \underline{39.8} & \textbf{41.7} & 35.9 & 38.8 & \underline{42.7} & \textbf{50.5} & \underline{62.1} & \textbf{64.1} & 59.2 & 57.3 & 68.0 & 68.9 & \textbf{72.8} & \underline{71.8} \\
BC-200 & 5.0 & 5.5 & \underline{6.0} & \textbf{7.0} & 5.5 & 7.0 & \underline{9.0} & \textbf{9.5} & 24.5 & \textbf{27.5} & \underline{27.0} & 22.0 & 35.0 & 34.0 & \underline{36.5} & \textbf{39.0} \\
SEAL-0 & 10.8 & 10.8 & \underline{15.3} & \textbf{16.2} & 22.5 & \underline{24.3} & 23.4 & \textbf{26.1} & \underline{31.5} & 29.7 & \underline{31.5} & \textbf{34.2} & 34.2 & 35.1 & \underline{36.0} & \textbf{36.9} \\
XBench-DS & \underline{41.0} & 39.0 & \textbf{43.0} & \textbf{43.0} & 47.0 & 50.0 & \underline{51.0} & \textbf{60.0} & 66.0 & \underline{69.0} & \textbf{71.0} & \underline{69.0} & \underline{71.0} & 68.0 & \underline{71.0} & \textbf{72.0} \\
BC-ZH & 8.0 & \underline{9.3} & 9.0 & \textbf{9.7} & 15.2 & \underline{15.9} & \underline{15.9} & \textbf{16.3} & 36.0 & \underline{41.5} & 41.0 & \textbf{43.3} & 40.5 & 39.8 & \underline{41.9} & \textbf{43.6} \\
\bottomrule
\end{tabular*}

\smallskip
\begin{tabular*}{\linewidth}{@{\extracolsep{\fill}}l*{4}{c}@{\hspace{8pt}}*{4}{c}@{\hspace{8pt}}*{4}{c}@{\hspace{8pt}}*{4}{c}@{}}
 & \multicolumn{4}{c}{Qwen3.5-397B-A17B} & \multicolumn{4}{c}{Qwen3.6-35B-A3B} & \multicolumn{4}{c}{Gemma-4-e2b-it} & \multicolumn{4}{c}{Gemma-4-e4b-it} \\
\cmidrule(lr){2-5}\cmidrule(lr){6-9}\cmidrule(lr){10-13}\cmidrule(lr){14-17}
Benchmark & $k{=}4$ & $k{=}8$ & $k{=}16$ & $k{=}32$ & $k{=}4$ & $k{=}8$ & $k{=}16$ & $k{=}32$ & $k{=}4$ & $k{=}8$ & $k{=}16$ & $k{=}32$ & $k{=}4$ & $k{=}8$ & $k{=}16$ & $k{=}32$ \\
\midrule
GAIA & \textbf{76.7} & \textbf{76.7} & 71.8 & \underline{72.8} & 78.6 & \underline{79.6} & \textbf{80.6} & \underline{79.6} & \underline{30.1} & 27.2 & 28.2 & \textbf{31.1} & 24.3 & 30.1 & \underline{34.0} & \textbf{36.9} \\
BC-200 & 37.0 & 38.0 & \underline{42.0} & \textbf{42.5} & 36.0 & \textbf{41.5} & \underline{38.0} & 37.5 & \textbf{2.5} & 0.5 & \underline{1.0} & \underline{1.0} & 1.0 & \textbf{2.5} & \underline{1.5} & \underline{1.5} \\
SEAL-0 & \underline{37.8} & 36.0 & \underline{37.8} & \textbf{40.5} & 24.3 & \textbf{28.8} & \textbf{28.8} & \underline{27.0} & 4.5 & \textbf{9.0} & 4.5 & \underline{6.3} & \underline{5.4} & \textbf{7.2} & \underline{5.4} & \textbf{7.2} \\
XBench-DS & \underline{79.0} & 77.0 & \underline{79.0} & \textbf{80.0} & \underline{73.0} & \textbf{74.0} & \textbf{74.0} & \textbf{74.0} & 28.0 & \underline{30.0} & 28.0 & \textbf{31.0} & 27.0 & 26.0 & \underline{32.0} & \textbf{40.0} \\
BC-ZH & \underline{55.7} & 55.4 & \textbf{57.1} & 55.4 & 35.3 & \underline{39.8} & \textbf{41.5} & 38.8 & 1.7 & 2.4 & \underline{3.5} & \textbf{3.8} & \underline{3.8} & 2.8 & \textbf{4.5} & 3.5 \\
\bottomrule
\end{tabular*}
\vspace{-5pt}
\end{table*}

\begin{table}[t]
\centering
\small
\caption{Estimator--oracle agreement (\%): how often both select the same candidate action. Agents are Qwen3.5 variants.}
\label{tab:estimator-noise}
\vspace{-5pt}
\setlength{\tabcolsep}{6pt}
\renewcommand{\arraystretch}{0.95}
\begin{tabular*}{\linewidth}{@{\extracolsep{\fill}}llrrrr@{}}
\toprule
Model & Dataset & $k{=}4$ & $k{=}8$ & $k{=}16$ & $k{=}32$ \\
\midrule
397B & GAIA & 25.0 & 21.1 & 11.8 & 10.0 \\
122B & BC-200 & 40.9 & 21.4 & 17.6 & 22.5 \\
397B & XBench-DS & 30.8 & 7.9 & 0.0 & 0.0 \\
\bottomrule
\end{tabular*}
\end{table}

\label{sec:setup}

\phantomsection
\label{sec:lmtraining}

\textbf{Training Setup.}
Frozen agents collect real trajectories by sampling questions round-robin from DeepDive~\citep{deepdive}, MiroVerse~\citep{miroverse}, Nemotron RL~\citep{nemotronrl}, OpenSeeker~\citep{openseeker}, and REDSearcher~\citep{redsearcher}, excluding evaluation questions.
This yields 0.7M call--response pairs for the simulator and 1M training samples for the estimator.
All methods share two external tools:
\texttt{search} returns titles, links, and snippets, while \texttt{visit} retrieves pages and extracts goal-relevant evidence.
Extraction in \texttt{visit} is performed by a shared Qwen3.5-397B-A17B service, so its cost applies equally to all compared methods.
Both models are initialized from Qwen3.5-9B~\citep{qwen35} and use LoRA~\citep{lora} at a learning rate of $10^{-4}$, with simulator/estimator ranks of 64/32 and context limits of 16k/128k tokens.
The estimator learns discounted terminal correctness ($\gamma=0.95$) through a fully trainable HL-Gauss head~\citep{hlgauss} with 201 bins over $[-1,1]$ and a Gaussian standard deviation of 0.75 bin widths.
Both checkpoints are reused unchanged across agents.

\phantomsection
\label{sec:impl}
\textbf{Evaluation Setup.}
We report final-answer accuracy on GAIA~\citep{gaia}, BrowseComp-200 (BC-200)~\citep{browsecomp}, BrowseComp-ZH (BC-ZH)~\citep{browsecompzh}, SEAL-0~\citep{sealqa}, and XBench-DeepSearch (XBench-DS)~\citep{xbench}, following each benchmark's evaluation protocol.
The main comparison covers Qwen3.5-4B/9B/35B-A3B~\citep{qwen35}, Qwen3.6-35B-A3B~\citep{qwen36}, and Gemma-4-e2b-it/e4b-it~\citep{gemma4}.
For each model, all methods share the frozen agent and tools and use the model's recommended sampling settings.
\emph{Base Agent} runs one ReAct-style trajectory~\citep{react}, whereas \emph{BoN@4} and \emph{AV@4} run four and select by normalized-answer majority voting~\citep{selfconsistency} and prompted outcome verification~\citep{av}, respectively, without gold-answer access.
We also implement \emph{RAP@4}~\citep{rap}, but omit it from the main table after prompted simulation and self-evaluation yield little benefit on 4B/9B agents, as illustrated in Table~\ref{tab:rap}.
Each trajectory is limited to 100 committed agent rounds, 128k input tokens, and 9000\,s, with a direct-answer attempt upon reaching a limit.
Speculative proposals do not count toward the round limit.
Seer uses depth-one imagined search with candidate width $k\in\{4,8,16,32\}$, reporting $k=4$ and $k=32$ as low-width and scaled settings in the main comparison.
These widths are not compute-matched to four full trajectories, whose real tool calls are all incurred, while Seer executes only selected actions.
In tables, Base, BoN, and AV abbreviate Base Agent, BoN@4, and AV@4, respectively; $S_k$ denotes Seer with candidate width $k$, labeled Seer@$k$ in figures.
In all result tables, bold and underlined values indicate the best and second-best results within each comparison group, respectively, including ties.

\subsection{Analysis}
\label{sec:analysis}

\textbf{Main Results Across Agent Models}:
\label{sec:main}
Table~\ref{tab:main} shows that Seer supports Qwen and Gemma agents, reusing the same simulator and estimator without model-specific retraining.
With four candidates per decision ($S_4$), Seer outperforms both four-trajectory baselines on all five benchmarks with Qwen3.5-4B and reaches 47.0\% on XBench-DeepSearch with 9B, versus 38.0\% for both baselines.
Expanding to $S_{32}$ raises GAIA accuracy to 41.7\% and 50.5\% for 4B and 9B, respectively, and surpasses both baselines on every benchmark with 9B.
Gains extend to Gemma-4, with $S_{32}$ raising GAIA accuracy from 20.4\% to 31.1\% for e2b-it and from 21.4\% to 36.9\% for e4b-it.
Across all six models, $S_{32}$ improves over Base Agent in 26 of 30 comparisons and reaches 79.6\% on GAIA with Qwen3.6-35B-A3B without retraining either shared component.
Unlike trajectory-level selection, both Seer settings evaluate alternatives without real tool calls for discarded candidates, trading extra inference computation for broader search rather than paying to execute every alternative.

\textbf{Scaling to Larger Agent Models}:
\label{sec:model-scale}
Although designed for small agents, Seer also benefits substantially larger models (Table~\ref{tab:large-models}).
Reusing the same simulator and estimator without retraining, $S_{32}$ improves over Base Agent on all five benchmarks for both Qwen3.5-122B-A10B and Qwen3.5-397B-A17B and outperforms both four-trajectory baselines in nine of ten comparisons.
For 397B, it raises BrowseComp-200 accuracy from 34.5\% to 42.5\% and SEAL-0 from 31.5\% to 40.5\%.
These gains show that Seer's shared guidance transfers across model scales rather than benefiting only small agents, without implying these larger models are suitable for on-device deployment.

\textbf{Contributions of Prediction and Evaluation}:
\label{sec:ablation}
Table~\ref{tab:ablation} cumulatively builds Seer from the frozen Base Agent with four candidates per decision, all executed and ranked by prompted scoring.
Replacing prompted scoring with the estimator improves accuracy by 0.9--22.3 points on seven of eight models, except Qwen3.5-397B-A17B.
Retaining the estimator and adding simulator predictions lowers accuracy on six models at $k=4$ but eliminates real calls for discarded candidates.
Increasing Seer to $k=32$ matches or exceeds real-response selection at $k=4$ on six of eight models (Tables~\ref{tab:main}, \ref{tab:large-models}, and \ref{tab:ablation}), showing that broader search can compensate for prediction-related losses through additional inference rather than execution of discarded candidates.
Direct scoring without outcome simulation trails Seer on seven of eight models with one tie, echoing WebDreamer~\citep{webdreamer}: an action's value hinges on content invisible at decision time, such as the page behind a search result, which only simulated responses can reveal before selection.

\textbf{Accuracy versus Real Tool Use}:
\label{sec:cost}
Figure~\ref{fig:tool-cost} shows that Seer improves the accuracy--tool-cost trade-off on BrowseComp-200 across all agents.
Its tool-call costs remain close to those of Base Agent across the tested widths, whereas BoN@4 and AV@4 incur substantially more real calls by executing four complete trajectories.
These results illustrate the benefit of evaluating alternatives through simulation and executing only selected actions: additional inference computation enables more effective search without paying the real tool costs of discarded candidates.

\textbf{Scaling Candidate Width}:
\label{sec:ksweep}
Across all agents and benchmarks (Table~\ref{tab:ksweep}), $k=32$ falls below the best tested width in 14 of 40 settings, although it achieves the highest accuracy, including ties, on all five benchmarks for Qwen3.5-4B/9B.
Optimal widths depend on both agent and task: Qwen3.6-35B-A3B peaks at $k=16$ on GAIA and $k=8$ on BrowseComp-200, whereas Gemma-4-e2b-it favors $k=32$ and $k=4$, respectively.
Table~\ref{tab:estimator-noise} further probes ranking reliability by comparing the estimator's action choice with that of a GPT-5.5 oracle given the ground-truth answer for this diagnostic.
Agreement generally decreases with $k$, though not monotonically on BC-200, consistent with greater sensitivity to estimator noise when selecting among more candidates.
Together, these results suggest a trade-off between candidate coverage under simulator noise and ranking reliability under estimator noise.
Wider search offers more opportunities to find useful actions despite imperfect predictions, without reducing per-call simulator noise, but also increases the risk of selecting overvalued branches, potentially compounded by such errors.
This trade-off motivates selecting a specific candidate width rather than simply maximizing $k$.

\section{Conclusion}

We presented Seer, an inference-time framework for local guidance in deep research that separates agent proposals from learned tool-response prediction and value estimation.
Trained with complementary supervision from a shared trajectory corpus, its simulator and estimator guide different agent models without retraining, using imagined search but only real responses as evidence.
Experiments show substantial small-agent gains and transfer across model families and scales, with tool-call costs close to those of Base Agent without search guidance in the measured comparisons.
We believe Seer suggests a path toward capable, locally controlled research agents through reusable guidance rather than model-specific adaptation or costly external exploration.

\section*{Acknowledgments}

This work was supported by the National Natural Science Foundation of China under Grant No. 625B2122 (Youth Student Basic Research Program for PhD Students) and Grant No. 62427820 (National Major Scientific Instrument and Equipment Development Project).

\section*{LLM Usage Statement}

LLM assistance was used for language editing, consistency checks, and manuscript presentation.
The authors retain full responsibility for reviewing the final manuscript and for all technical claims and conclusions.

{
    \small
    \bibliographystyle{ieeenat_fullname}
    \bibliography{refs}
}

\onecolumn
\appendix
\etocdepthtag.toc{appendix}
\begingroup
\etocsettagdepth{main}{none}
\etocsettagdepth{appendix}{section}
\etocstandardlines
\makeatletter
\renewcommand*{\l@section}{\@dottedtocline{1}{0em}{2em}}
\makeatother
\etocsettocstyle{
  \section*{Appendix Contents}
  \noindent\textbf{Section}\hfill\textbf{Page}\par\smallskip
}{}
\tableofcontents
\endgroup
\par\medskip

\section{Notation}
\label{app:notation}

Table~\ref{tab:notation} summarizes the notation used in this paper.
Note that we distinguish the frozen model's proposal distribution $\pi$ from the agent's search-guided action distribution $\pi_{\mathcal{G}}$, and real observations $o_t$ from predicted observations $\hat{o}^{(i)}$.
Lowercase $k$ denotes candidate width, whereas uppercase $K$ denotes the number of value bins.

\begin{table*}[t]
\centering
\footnotesize
\setlength{\tabcolsep}{3pt}
\caption{Notation used in the paper.}\label{tab:notation}
\begin{tabular}{@{}p{0.25\linewidth}p{0.71\linewidth}@{}}
\toprule
Symbol & Meaning \\
\midrule
\multicolumn{2}{@{}l}{\textbf{Research interactions and resource constraints}}\\*
$q,\mathcal{D}$ & Research request and task distribution, with $q\sim\mathcal{D}$. \\
$\mathcal{G},\pi_{\mathcal{G}}$ & Research agent and its action distribution, including search guidance. \\
$\pi$ & Action-proposal distribution of the frozen agent model. \\
$t,s_t$ & Real-round index and committed history, initially containing $q$ and agent instructions. \\
$a_t,\hat{a}$ & Selected action (tool use or final submission) and submitted answer, respectively. \\
$o_t,\varnothing$ & Real tool responses and the empty response for final submission. \\
$\mathcal{E},\oplus$ & Real execution and concatenation in $s_{t+1}=s_t\oplus(a_t,\mathcal{E}(a_t))$. \\
$\operatorname{calls}(a),c$ & Ordered calls in action $a$ and one call (tool name and arguments). \\
$\xi,R_{\mathrm{term}}(\xi;q)$ & Execution record and terminal quality score; binary correctness in the evaluated tasks, with zero for unanswered tasks. \\
$\mathbf{c}(\mathcal{G},\xi)$ & Resource vector $(c_{\mathrm{param}}(\mathcal{G}),c_{\mathrm{tool}}(\xi))^\top$. \\
$c_{\mathrm{param}}(\mathcal{G})$ & Total number of locally deployed model parameters, including those of the agent model and both decision models in Seer. \\
$c_{\mathrm{tool}}(\xi)$ & All real calls in $\xi$, including those for discarded candidates if executed; batched calls count individually. Seer executes only selected actions. \\
$\mathbf{B},\preceq$ & Resource-budget vector and componentwise inequality. \\
\midrule
\multicolumn{2}{@{}l}{\textbf{Tool-response prediction}}\\*
$W_\phi,\phi$ & Tool-response simulator and its trainable parameters. \\
$\widehat{\mathcal{E}}$ & Local response approximation, collecting predictions in call order. \\
$\hat{o}_c$ & Predicted response to an individual call, sampled from $W_\phi(\cdot\mid c)$. \\
$\mathcal{D}_W,N_W$ & Dataset of retained call--response pairs and its size. \\
$c_i,o_i$ & Tool call and observed response in simulator training pair $i$. \\
$\ell,o_\ell,o_{<\ell},|o|$ & Response-token index, current token, preceding tokens, and response length. \\
$\mathcal{L}_W(\phi)$ & Response-only negative log-likelihood used to train the simulator. \\
\midrule
\multicolumn{2}{@{}l}{\textbf{Outcome-supervised value estimation}}\\*
$V_\theta,\theta$ & Value estimator over histories and its trainable parameters, including its head. \\
$\mathcal{D}_V$ & Dataset of history prefixes and discounted terminal correctness targets. \\
$q_\xi,s_{\xi,t},T_\xi$ & Question, prefix at position $t$, and terminal history index of $\xi$. \\
$\gamma,G_{\xi,t}$ & Discount factor and target $\gamma^{T_\xi-t}R_{\mathrm{term}}(\xi;q_\xi)$, discounted by remaining interaction steps, not tokens or individual tool calls. \\
$s,G$ & Generic history and scalar target in a value-training pair. \\
$K,b$ & Number of uniform value bins and bin index $b\in\{1,\ldots,K\}$. \\
$\beta_0,\ldots,\beta_K$ & Ordered bin boundaries delimiting the value-head support. \\
$z_b$ & Bin center $(\beta_{b-1}+\beta_b)/2$. \\
$\sigma,\Phi$ & Smoothing standard deviation and standard normal cumulative distribution function. \\
$y_b(G),p_\theta(b\mid s)$ & Normalized Gaussian target mass and predicted probability for bin $b$. \\
$\mathcal{L}_V(\theta)$ & Soft-label cross-entropy; scores are decoded as $V_\theta(s)=\sum_b z_b p_\theta(b\mid s)$. \\
\midrule
\multicolumn{2}{@{}l}{\textbf{Imagined search and commitment}}\\*
$\bar{s},\hat{s}^{(i)}$ & Selected-node history and imagined history for candidate $i$. \\
$a^{(i)},\hat{o}^{(i)},v^{(i)}$ & Candidate action, predicted responses, and estimated history value. \\
$k,m$ & Candidate width (maximum proposals per expansion) and actual number of valid proposals, with $1\le m\le k$. \\
$N_{\mathrm{sim}},n,d_{\max}$ & Search simulations per real round, search-simulation index, and depth cap. \\
$u,e,e^{(i)}$ & Parent node, outgoing edge, and edge corresponding to candidate $i$. \\
$P(e),c_{\mathrm{puct}}$ & Edge prior ($1/m$ for new siblings) and PUCT exploration coefficient. \\
$N(u),N(e),Q(e)$ & Node visits, edge visits, and mean value backed up through an edge. \\
$\operatorname{score}(e),v$ & PUCT selection score and simulation-endpoint value. \\
$e^*,\operatorname{action}(e^*)$ & Most-visited root edge selected for commitment and its action. \\
\bottomrule
\end{tabular}
\end{table*}

\section{Extended Background and Related Work}
\label{app:related-work}

We begin with the research-agent setting and its resource constraints, then expand the three directions reviewed in \S\ref{sec:related} with additional methods and details not covered in the main text.
Efficient research agents address the cost and capability of the deployed model, while inference-time guidance improves how its proposals are selected.
Within inference-time guidance, model-based lookahead addresses a further question: how to evaluate candidate consequences without executing every alternative.
These directions are complementary, not mutually exclusive.

\subsection{Research Agents and Tool Use}

Research agents combine a language model's parametric knowledge with information acquired through tools.
WebGPT~\citep{webgpt} combines browser-assisted question answering with human-feedback training, while ReAct~\citep{react} interleaves reasoning and actions so that subsequent decisions incorporate environmental observations.
Toolformer~\citep{toolformer} learns when and how to invoke APIs through self-supervised examples.
Together, these approaches establish the interaction pattern underlying research agents: selecting tools, observing their outputs, and using the resulting information to guide subsequent reasoning.
InternAgent-1.5~\citep{feng2026internagent} extends this pattern to scientific discovery by coordinating generation, verification, and evolution across computational and empirical tasks.

Within this setting, the task determines which observations are useful.
WebArena~\citep{webarena} evaluates functional task completion in interactive websites, whereas the benchmarks used in this study assess answers assembled through search and page retrieval (\S\ref{sec:setup}).
Research agents must therefore decide which query or goal-conditioned page visit is likely to reveal useful information.
This creates two deployment concerns: the local resources needed to run the model and the external tool calls needed to gather evidence.
The following subsections consider improving the deployed model first, then guiding its decisions without updating the deployed model.

\subsection{Efficient Research Agents}

One approach is to reduce the cost of the deployed model or improve its capabilities through training.
AWQ~\citep{awq} reduces weight storage and inference costs through activation-aware quantization.
MiniLLM~\citep{minillm} distills larger language models into smaller ones, while generalized knowledge distillation~\citep{gkd} uses teacher feedback on student-generated sequences to address the mismatch between training and inference.
These methods make model compression and capability transfer central to efficient deployment.
Beyond language models, STORM~\citep{lv2026spatial} reduces visual tokens while preserving spatial structure in visual state space models, whereas Ferret~\citep{zhou2025ferret} adapts model partitioning and pipeline execution to changing memory budgets during online continual learning.
Both illustrate how efficiency also depends on the representation and execution of computation, complementing methods that improve an agent's decisions.

Reinforcement learning further targets the agent's tool-use behavior.
Search-R1~\citep{searchr1} trains models to interleave reasoning with retrieval using outcome-based rewards, and DeepResearcher~\citep{deepresearcher} optimizes research agents through real web interactions.
AgentCPM-Explore~\citep{agentcpmexplore} extends this direction to long-horizon research with an edge-scale model.
Alongside reinforcement learning, Agents-A1~\citep{bai2026scaling} combines supervised fine-tuning with distillation from domain-specific teachers to strengthen long-horizon capabilities in a 35B model.
Although such training can strengthen the proposal model, adapting another backbone requires additional optimization.
This motivates the complementary approach considered next: guiding proposals at inference time while keeping the deployed agent frozen.
Such guidance requires no retraining of the deployed agent, and reusing the same guidance components across agents avoids repeated model-specific adaptation.

Reuse across deployments is also studied in federated learning, although the objects being transferred differ.
HyperFedZero~\citep{zhou2026deploying} generates models for previously non-participating clients without client-side fine-tuning, while ForgeVLA~\citep{zhou2026forgevla} learns vision-language-action models from distributed robot experience without centralizing raw data or requiring manual language annotations.
These approaches address model deployment and collaborative training across clients, whereas Seer reuses prediction and evaluation models to guide frozen agents at inference time.

\subsection{Inference-Time Guidance}

Inference-time guidance can act on completed outputs or intervene at intermediate decisions.
This distinction matters for research agents because generating a candidate answer may already require substantial tool use.

\textbf{Output-level refinement.}
Learned verifiers~\citep{verifiers} score candidate solutions, self-consistency~\citep{selfconsistency} aggregates independently sampled reasoning paths, and Self-Refine~\citep{selfrefine} alternates generation with feedback and revision.
For deep research, asymmetric verification~\citep{av} allocates a relatively modest verification budget to selecting among expensive search trajectories, while DeepVerifier~\citep{deepverifier} uses test-time rubric-guided verification in an iterative refinement loop.
These methods can select an existing answer or inform a subsequent revision, but cannot recover tool calls already spent producing the candidates.
For example, the BoN@4 and AV@4 baselines used in our experiments execute all four trajectories before selecting an answer (\S\ref{sec:setup}).
This cost motivates guidance before the next real action, rather than only after a trajectory has been completed.

\textbf{Decision-level evaluation.}
Tree of Thoughts~\citep{tot} explores intermediate reasoning states through search and self-evaluation, while Reflexion~\citep{reflexion} uses verbal feedback from earlier attempts to guide later behavior.
For interactive agents, LATS~\citep{lats} combines tree search, value estimation, and reflection with environmental feedback, and \citet{treeweb} perform search in the actual web environment.

\textbf{Learning reusable evaluators.}
Search determines which alternatives to explore, while evaluation determines which appear promising.
Rather than relying only on prompted self-evaluation, a separately trained evaluator can provide this ranking signal.
Step-level process supervision~\citep{stepverify} trains evaluators from intermediate feedback, and OmegaPRM~\citep{omegaprm} automates process-supervision collection through Monte Carlo search for mathematical reasoning.
AgentPRM~\citep{agentprm} uses Monte Carlo reward targets in an actor--critic framework to train agent policies, rather than restricting guidance to inference time.
AgentRM~\citep{agentrm} learns a reward model for test-time guidance that transfers across tasks and policy models, while RAG-Gym~\citep{raggym} studies transferable search critics for information seeking.

These methods separate the model that proposes actions from the component that evaluates them.
However, changing the evaluator does not by itself remove the cost of obtaining candidate observations.
Branches explored through real interaction still incur tool calls even when discarded.
This remaining cost motivates model-based lookahead, which supplies predicted observations for evaluation before execution.

\subsection{Model-Based Lookahead}

Model-based lookahead addresses how candidate consequences are obtained.
Instead of executing every alternative, an agent can use a learned or prompted world model to predict aspects of the resulting observation or state.
The distinction between using such predictions for policy learning and for inference-time planning clarifies Seer's relationship to this literature.

\textbf{Foundations of world modeling.}
Dyna~\citep{dyna} integrates real experience with model-based planning, while \citet{worldmodels} learn compact representations of environment dynamics and train a controller within a learned model.
MuZero~\citep{muzero} learns latent dynamics together with planning-relevant reward, policy, and value predictions.
DreamerV3~\citep{dreamerv3} learns behavior from imagined trajectories, and Model-Based Policy Optimization~\citep{mbpo} uses short model-generated rollouts branched from real data to limit the effects of model bias.
These approaches establish the use of internal simulation, with different choices of predictive representation and learning objective.

\textbf{World models for agent training.}
Language-based and synthetic environments extend this idea to learning interactive policies.
Dynalang~\citep{dynalang} learns a multimodal world model that predicts future text and image representations and trains a policy from imagined rollouts.
DreamGym~\citep{dreamgym} generates transitions and feedback for agent reinforcement learning, and DynaWeb~\citep{dynaweb} combines rollouts in a learned web environment with real expert trajectories.
Qwen-AgentWorld~\citep{qwenagentworld} learns language world models across multiple agent environments and studies their use both as simulators for reinforcement learning and as a foundation for agent training.
In these settings, predicted experience supports learning the agent itself.

\textbf{Language world models for inference-time planning.}
Predictions can also guide a fixed agent without updating its policy.
RAP~\citep{rap} repurposes an LLM as both a reasoning agent and a world model, combining predicted transitions with Monte Carlo tree search and task-dependent rewards.
For web navigation, WMA~\citep{wma} learns transition-focused natural-language descriptions of changes between observations rather than reproducing full HTML states.
WebDreamer~\citep{webdreamer} combines simulated action outcomes with value-based selection and trains specialized world models using synthesized interaction data.
SiRA~\citep{sira} uses natural-language belief states for goal-oriented simulative planning, while MCP-Cosmos~\citep{mcpcosmos} integrates world models with MCP tools to support planning before execution.
These methods bring language-based prediction into candidate comparison rather than requiring real interaction for every alternative.

\textbf{Relation to Seer.}
Seer combines model-based lookahead with a reusable evaluator for frozen research agents.
Its simulator can be viewed as a specialized language world model that predicts tool responses from individual calls, rather than modeling the full external environment.
The estimator separately scores the resulting histories using discounted terminal correctness from real continuations.
These targets reflect observed outcomes and remaining interaction steps, not direct judgments of each action's correctness.
Both components are trained offline on real trajectories and reused across agents, rather than updating the agent through imagined experience.
This combines call-level prediction with history-level evaluation while keeping action proposal separate.
Only selected actions are executed, and only their real responses enter the next committed history.
Seer therefore complements model compression and policy training while moving candidate comparison before real tool execution.

\begin{algorithm}[htp]
\caption{Seer: imagined search and real execution in one round}
\label{alg:seer-search}
\small
\begin{algorithmic}[1]
\REQUIRE nonterminal committed history $s_t$; frozen proposal distribution $\pi$;
         models $W_\phi,V_\theta$; $N_{\mathrm{sim}},k,d_{\max}\ge1$;
         $c_{\mathrm{puct}}\ge0$
\STATE Initialize root $s_t$ at depth zero; new nodes have $N=0$
       and new edges have $N=Q=0$
\FOR{$n=1$ \TO $N_{\mathrm{sim}}$}
  \STATE Select a path to a terminal, unexpanded, or depth-capped node
         with history $\bar{s}$ using Eq.~(\ref{eq:seer-puct})
  \IF{the node is nonterminal, unexpanded, and below $d_{\max}$}
    \STATE Independently sample $m$ valid actions $\{a^{(i)}\}_{i=1}^{m}$
           from $\pi(\cdot\mid\bar{s})$, with $1\le m\le k$
    \STATE Batch their tool calls through $W_\phi$ to obtain $\hat{o}^{(i)}$
           in call order; use $\varnothing$ for final-answer candidates
    \STATE Add children $\hat{s}^{(i)}=\bar{s}\oplus(a^{(i)},\hat{o}^{(i)})$
           with edge priors $P(e^{(i)})=1/m$; mark final-answer children terminal
    \STATE Extend the path through the first new edge
  \ENDIF
  \STATE $v\leftarrow V_\theta(\text{history at path endpoint})$,
         including terminal histories
  \STATE For each path edge: $N(e)\leftarrow N(e)+1$,
         $Q(e)\leftarrow Q(e)+(v-Q(e))/N(e)$
  \STATE Increment visit counts of all path nodes, including the root
\ENDFOR
\STATE Select $a_t$ by Eq.~(\ref{eq:seer-commit})
\IF{$a_t$ is a final-answer submission}
  \STATE Return the answer and terminate
\ELSE
  \STATE Execute only $a_t$'s tool calls to obtain $o_t=\mathcal{E}(a_t)$
  \STATE Discard the imagined tree; return the real history
         $s_{t+1}=s_t\oplus(a_t,o_t)$ for the next round
\ENDIF
\end{algorithmic}
\end{algorithm}

\section{Seer Search Algorithm}
\label{app:planner}

Algorithm~\ref{alg:seer-search} describes one round of imagined search followed by real execution, as introduced in \S\ref{sec:planner}.
Search begins at the committed history and explores branches with separate imagined continuations, up to a depth cap $d_{\max}$.
During each simulation, selection follows Eq.~(\ref{eq:seer-puct}), and the estimator scores the endpoint to assess its potential for task completion.
This value is then backed up along the traversed path, updating the mean value of each edge.
After search, the action on the most visited edge from the root is selected for real execution according to Eq.~(\ref{eq:seer-commit}).

Note that no gold-answer judgment is queried during search, including for terminal candidates.
After a tool-using action is selected, its actual responses replace the predicted continuation rather than being appended to it.
Thus, each subsequent round starts from a history containing only committed actions and real tool observations.

\section{Tool-Response Simulator Implementation Details}
\label{app:world}

The simulator $W_\phi$ approximates the information returned by the shared tools, rather than the full state of the web.
Its modeling unit is an individual call $c$: a tool name together with its arguments.
As in Eq.~(\ref{eq:seer-simulation}), predictions for a multi-call action are assembled in the original call order.
This factorization makes response prediction independent of the agent architecture while preserving the format expected by the value estimator.

\subsection{Supervision from Observed Tool Responses}

The real trajectories collected in \S\ref{sec:training} provide call--response supervision for both \texttt{search} and \texttt{visit}.
Search outputs contain titles, links, and snippets; visit outputs contain page content extracted for the specified retrieval goal.
We normalize equivalent call representations, remove infrastructure exceptions, and deduplicate repeated call--response pairs.
Empty results and inaccessible pages are represented consistently, while substantive returned content is retained.
These steps reduce variation caused by the collection infrastructure without filtering examples according to whether the final answer was correct.
An unsuccessful research trajectory can still supply a valid observation of what a tool returns.

The resulting simulator corpus contains 0.7M pairs, as reported in \S\ref{sec:setup}.
Validation is separated by normalized call identity so that repeated observations of the same call do not appear on both sides of the split.
This differs from the estimator's question-disjoint split: the simulator learns a call-level conditional distribution, whereas the estimator learns to assess question-dependent histories.

\subsection{Tool-Conditioned Prediction}

The simulator uses the same tool-specific input templates during training and inference.
The boxed templates below make the conditioning information explicit.
The query, URL, and goal are supplied by the proposed tool call; the enclosing conversation and the identity of the proposing model are not additional inputs.
For a page visit, conditioning on the goal lets the simulator predict a focused extraction rather than an undirected reconstruction of the page.

\begin{promptbox}{Search simulator prompt}
Act as a web search engine simulator. Return the top search results for the query below.\par\medskip
Query: \textcolor{promptaccent}{\{query\}}
\end{promptbox}

\begin{promptbox}{Page-visit simulator prompt}
Act as a webpage extraction tool simulator. Extract content from the URL that is relevant to the goal.\par\medskip
URL: \textcolor{promptaccent}{\{url\}}\par
Goal: \textcolor{promptaccent}{\{goal\}}
\end{promptbox}

Training minimizes the response-only negative log-likelihood in \S\ref{sec:world}.
The instruction and call arguments supply context but do not contribute target tokens to the loss.
This objective learns the conditional form and content of tool responses, rather than training the simulator to choose actions or judge whether a research question has been solved.

\subsection{Configuration and Use in Search}

Following \S\ref{sec:setup}, we initialize the simulator from Qwen3.5-9B and train it with LoRA rank 64, a learning rate of $10^{-4}$, and a context limit of 16k tokens.
Each training example must retain response tokens within this limit, so examples that leave no tokens for response supervision are excluded.
After training, the same checkpoint is reused across agent models without further updates.

At inference, candidate calls are grouped for simulation, and each predicted response is attached only to its corresponding imagined branch.
Proposals that submit a final answer contain no tool calls and therefore require no simulator output.
These predictions support candidate comparison but do not serve as evidence.
Once an action is selected, its tool calls are executed in the real environment, and only the resulting observations enter the next root history.
This procedure allows comparison before execution while keeping subsequent reasoning grounded in real tool responses.

\section{Value Estimator Implementation Details}
\label{app:value}

The estimator $V_\theta$ complements call-level prediction by assessing a history in the context of its research question.
Its supervision associates intermediate histories with observed task outcomes and their distance from completion.
It provides an outcome-supervised scalar ranking signal for candidate histories, rather than predicting tool responses or generating critiques.

\subsection{History-Level Targets}

We retain deduplicated trajectories with submitted answers and exclude unanswered runs that end in errors or timeouts, as in \S\ref{sec:value}.
A judge compares each submitted answer with its gold answer to assign binary terminal correctness.
The gold answer and judge output are used only to construct labels; neither is appended to the estimator's input history.
For a retained trajectory $\xi$, each labeled prefix receives
\begin{equation}
G_{\xi,t}=\gamma^{T_\xi-t}R_{\mathrm{term}}(\xi;q_\xi),
\qquad \gamma=0.95.
\end{equation}
The distance $T_\xi-t$ counts remaining interaction steps, not tokens or individual calls within an action.
Correct trajectories have positive targets that approach one at completion, whereas incorrect submitted answers induce zero targets throughout their trajectories.
Excluding unfinished runs from this training set does not change the evaluation convention that unanswered questions receive a terminal quality score of zero.

Discounting distinguishes prefixes associated with successful completion at different remaining distances without requiring manual step-by-step annotation.
However, each target describes an observed continuation from the shared collection corpus; it does not assert that the immediately preceding action was individually correct.
Since the corpus contains continuations from different frozen models, the learned score reflects this supervision rather than representing a calibrated success probability for a particular deployment policy.
These semantics are preserved when the estimator ranks imagined histories at inference.

\subsection{Gaussian-Smoothed Value Classification}

We replace the language-modeling head with a fully trainable HL-Gauss classification head~\citep{hlgauss}.
As specified in \S\ref{sec:setup}, the head has $K=201$ uniform bins over $[-1,1]$.
Their width is $\beta_b-\beta_{b-1}=2/201$, and the smoothing standard deviation is $\sigma=0.75(2/201)$.
Although the supervision targets lie in $[0,1]$, the classification support is the wider interval $[-1,1]$; the support should not be confused with the range of observed labels.

For each target, Eq.~(\ref{eq:seer-hlgauss}) integrates a Gaussian over the bins and normalizes its mass over the bounded support.
This gives neighboring bins similar supervision for nearby target values, rather than treating two targets on opposite sides of a bin boundary as unrelated classes.
Training minimizes soft-label cross-entropy, and inference returns the expectation of bin centers under the predicted distribution, as in \S\ref{sec:value}.
The scalar expectation is the value used for search and backup; no text-generation prompt or gold-answer comparison is required to obtain it.

\subsection{Causal Prefix Training and Shared Deployment}

The estimator corpus contains 1M training samples derived from the real trajectories (\S\ref{sec:setup}).
We split training and validation by question, keeping every trajectory and labeled prefix for the same question together.
This prevents closely related continuations from leaking across the split.
Each trajectory is processed in one causal forward pass, with supervision read at the endpoints of its labeled prefixes.
An endpoint representation can attend only to its own history, so later actions and the final submission are unavailable when scoring an earlier prefix.

Shared-prefix training requires each shorter history to have exactly the same tokens as the corresponding prefix of the full trajectory.
We exclude trajectories beyond the context limit rather than changing the evidence available to earlier prefixes through truncation.
These conditions preserve the intended history--target pairing while avoiding the repeated encoding of shared context.

The estimator is initialized from Qwen3.5-9B and uses LoRA rank 32, a learning rate of $10^{-4}$, and a 128k-token context limit, together with the fully trainable classification head.
Its checkpoint is reused unchanged across agents.
At deployment, it scores the history formed by appending a candidate action and its predicted responses; final-answer candidates are scored with an empty response.
The experiments use depth-one imagined search, so each candidate extends a real root by only one imagined action before evaluation.

\section{Simulator Case Studies: Predicted and Real Tool Responses}
\label{app:simulator-cases}

We compare predictions from the tool-response simulator $W_\phi$ with the corresponding real tool responses.
Unlike the trajectory comparisons in Appendix~\ref{app:cases}, these cases examine the content supplied to the value estimator, not final-answer accuracy or realized tool savings.
The shared interface has two tools: \texttt{search} returns result lists, and \texttt{visit} retrieves pages and extracts goal-relevant content (\S\ref{sec:setup}).
Page reading is therefore covered by \texttt{visit}, not a separate \texttt{read} tool, for which no examples are available here.

The boxes distinguish call arguments, predicted responses, and real responses.
Queries and outputs are condensed where indicated, and descriptive summaries are not verbatim tool logs.
Any research context is provided for the reader: $W_\phi$ receives only the call and its tool-specific instruction, not the enclosing conversation (Appendix~\ref{app:world}).
We examine agreement and discrepancies in response content rather than treating textual similarity as a measure of factual correctness.
These comparisons provide a qualitative analysis rather than an estimate of average prediction accuracy.

\subsection{Predicting Search Results}

\begin{promptbox}[breakable,fontupper=\small]{Simulator Case A: A specific historical search result}
\textbf{Call.} \texttt{search}, with query:
\texttt{hatpin "If Men Use Their Bare Fists to Fight Life's Battles" 1919 advertisement}.
The investigation seeks the wording of an advertisement for the 1919 film \emph{Bare Fists}.
\par\medskip
\textbf{Predicted response.}
The result list contains \emph{The Hat-Pin: Fashionably Dangerous}, with URL
\url{https://femmefashionforward.wordpress.com/2020/09/28/the-hat-pin-fashionably-dangerous/}.
Its snippet connects the film advertisement to the phrase ``If Men Use Their Bare Fists to Fight Life's Battles''.
\par\medskip
\textbf{Real response.}
The returned hit has the same title, URL, and answer-relevant snippet linking the phrase to the 1919 advertisement.
\par\medskip
\textbf{Interpretation.}
The match preserves a concrete retrieval target and the relation among the film, date, and wording, rather than merely predicting a historical topic.
This is the information an imagined continuation can expose to $V_\theta$ when comparing candidate actions.
It does not itself verify the phrase in the committed history: a selected search must be executed, and its real response supplies \mbox{the actual evidence.}
\end{promptbox}

\begin{promptbox}[breakable,fontupper=\small]{Simulator Case B: Preserving useful results for noisy queries}
\textbf{Call (abbreviated).} \texttt{search}, with a long query combining
\texttt{"Different Tempo" rooster hamster animation distinct song versions composer} and additional terms.
The research goal is to connect animated song performances to their musical source.
The displayed query preserves the main terms, not the full original search argument.
\par\medskip
\textbf{Predicted response (summary).}
A relevant hit links the hamster performance to \emph{The Hampsterdance Song} and the melody \emph{Whistle-Stop} from Disney's 1973 \emph{Robin Hood}.
\par\medskip
\textbf{Real response (summary).}
The corresponding hit supports the same musical connection rather than redirecting the investigation to a different dominant topic.
\par\medskip
\textbf{Interpretation.}
The supplied response pair preserves a useful association despite distracting query terms.
Unlike Case A, the available excerpt does not reproduce an exact result title or URL, so the visible comparison supports agreement on the research direction rather than an independently checkable, character-level match.
It does not show that the simulator corrected the query or recovered the final composer answer in this case.
\end{promptbox}

\begin{promptbox}[breakable,fontupper=\small]{Simulator Case C: Matched timeout, not invented content}
\textbf{Call (summary).} \texttt{search}, with an oversized query about the 2024 price range of Shanghai Gold Exchange Au(T+D), expanded with financial and unrelated technical terms.
\par\medskip
\textbf{Predicted response (summary).}
A Google search timeout for the expanded query, rather than a list of market results.
\par\medskip
\textbf{Real response (summary).}
The actual request also returns a Google search timeout.
\par\medskip
\textbf{Interpretation.}
Both responses indicate an absence of retrieved evidence, not a fact about gold prices.
Such a prediction could inform candidate-history evaluation, but this pair neither establishes that query length caused the timeout nor demonstrates reliable prediction of transient tool failures.
The training pipeline excludes infrastructure failures (\S\ref{sec:world}), so this diagnostic example should not be read as evidence that timeout messages were retained as training targets.
No downstream reformulation or avoided call is established by the response pair alone.
\end{promptbox}

\subsection{Predicting Goal-Conditioned Page Extractions}

\begin{promptbox}[breakable,fontupper=\small]{Simulator Case D: Preserving a biographical fact}
\textbf{Call.} \texttt{visit}, with URL
\url{https://www.dailymail.com/news/article-9925789/Elon-Musks-secret-sister-tells-Errol-Musk-turned-asked-DNA-test.html}.
\textbf{Goal:} Find how old Riette Nel's mother was when she gave birth to her.
\par\medskip
\textbf{Predicted response (condensed).}
Riette Nel was born as Mariette van Deventer in Ladysmith, South Africa, on August 8, 1975.
Her birth certificate did not name a father, and her mother was 21 at her birth.
The predicted summary answers the goal with \emph{21 years old}.
\par\medskip
\textbf{Real response (condensed).}
The extraction reports the same birth name, date, place, and birth-certificate detail, and states that her mother was 21.
\par\medskip
\textbf{Interpretation.}
The response pair agrees on both the requested age and the context identifying whose mother and birth are being discussed.
This illustrates goal-conditioned prediction of an answer-relevant passage, not reconstruction of the entire article or independent validation of every claim it contains.
The predicted extraction remains part of an imagined history until the selected call obtains a real response.
\end{promptbox}

\begin{promptbox}[breakable,fontupper=\small]{Simulator Case E: Retaining an exact scientific award identifier}
\textbf{Call.} \texttt{visit}, with URL
\url{https://iopscience.iop.org/article/10.3847/2041-8213/acd54b}.
\textbf{Goal:} Find the NASA award number supporting work by R.\,G. Arendt.
\par\medskip
\textbf{Predicted response (condensed).}
Work by R.G.A. was supported by NASA under award number \texttt{80GSFC21M0002}.
The surrounding acknowledgement also identifies the South African Radio Astronomy Observatory as the operator of MeerKAT and a facility of the National Research Foundation.
The summary repeats \texttt{80GSFC21M0002} as the requested award number.
\par\medskip
\textbf{Real response (condensed).}
The extraction contains the identical award string and the same surrounding acknowledgement context.
\par\medskip
\textbf{Interpretation.}
Here, usefulness depends on preserving an exact alphanumeric identifier and its association with the author, not merely recognizing that the paper acknowledges NASA support.
The agreement shows that the predicted extraction retains the requested field despite nearby institutional information.
\end{promptbox}

\begin{promptbox}[breakable,fontupper=\small]{Simulator Case F: Textual agreement does not guarantee layout fidelity}
\textbf{Call.} \texttt{visit}, with URL
\url{https://lindas-literary-home.com/my-literary-home/essays-poems-with-commentaries/audre-lordes-father-son-and-holy-ghost/}.
\textbf{Goal:} Retrieve the text of Audre Lorde's \emph{Father Son and Holy Ghost}, including indentation, to identify which stanza has indented lines.
\par\medskip
\textbf{Predicted response (summary).}
The extraction reproduces the poem's line sequence and includes leading spaces before lines such as ``Not that his judgment eyes'' and ``have been forgotten''.
The summary claims the text preserves the poem's structure and line breaks.
\par\medskip
\textbf{Real response (summary).}
The extraction contains the same line sequence but normalizes away the leading spaces, while describing the stanza and line structure as preserved.
\par\medskip
\textbf{Interpretation.}
The pair agrees closely on textual content but differs on the field required by this particular goal: indentation.
A statement that structure is preserved cannot substitute for an actual layout check.
Neither high textual similarity nor leading spaces in a prediction establish the original page's typography.
Layout-sensitive questions therefore require verification against a real source representation that retains formatting, rather than relying on the imagined extraction or on a whitespace-normalized response alone.
\end{promptbox}

\section{Case Studies: Action Selection and Real Tool Use}
\label{app:cases}

We examine six cases of Seer's search guidance and the resulting committed research trajectories.
The first four contrast a correct Seer answer with unsuccessful Base Agent, BoN@4, and AV@4 runs; the final two compare successful runs with different real tool costs.
All Seer runs use candidate width $k=32$ and depth cap $d_{\max}=1$.
Questions and trajectory excerpts are condensed for readability, rather than reproduced as verbatim tool logs.
These qualitative comparisons complement the aggregate results in \S\ref{sec:analysis}.

Table~\ref{tab:cases} summarizes the outcomes and resource use.
We count $c_{\mathrm{tool}}(\xi)$ as in \S\ref{sec:formulation}: every real \texttt{search} or \texttt{visit} call counts, including calls in discarded BoN@4 and AV@4 trajectories, and calls batched in one action count individually.
Final-answer submissions contribute no tool calls.
For Seer, $W_\phi$ predicts candidate tool responses and $V_\theta$ scores the resulting imagined histories before search selects an action for real execution.
These local computations are not real tool calls, but they are not free computation either.

\begin{table}[htbp]
\centering
\small
\caption{Case outcomes and real tool-call counts. C denotes a correct answer; I denotes an incorrect or unanswered result; --- indicates a comparison not included in the case report. Base, BoN, AV, and $S_{32}$ follow the notation in \S\ref{sec:setup}.}
\label{tab:cases}
\setlength{\tabcolsep}{5pt}
\begin{tabularx}{\linewidth}{@{}Xrrrr@{}}
\toprule
Case & Base & BoN & AV & $S_{32}$ \\
\midrule
1. OpenCV contributor & 11 (I) & 39 (I) & 50 (I) & \textbf{3 (C)} \\
2. Hiking location & 54 (I) & 178 (I) & 124 (I) & \textbf{17 (C)} \\
3. UN parliamentary seats & 40 (I) & 135 (I) & 104 (I) & \textbf{16 (C)} \\
4. NBA final-game margin & 14 (I) & 62 (I) & 47 (I) & \textbf{8 (C)} \\
\midrule
5. Encoder-layer comparison & 11 (C) & --- & --- & \textbf{2 (C)} \\
6. Elevator-accident article & --- & 21 (C) & --- & \textbf{2 (C)} \\
\bottomrule
\end{tabularx}
\end{table}

\subsection{Correct Answers from Focused Trajectories}

\begin{casebox}{Case 1: Turning a retrieved source into an identity check}
\textbf{Setting.} GAIA; Qwen3.5-9B.

\textbf{Question.} Which contributor to the OpenCV version that added Mask-RCNN support shares a Latin-transliterated name with a former Chinese head of government?
\textbf{Labeled answer:} Li Peng.

\textbf{Seer's committed trajectory: 3 calls (2 searches, 1 visit).}
A search for OpenCV's addition of Mask-RCNN support returns the ChangeLog.
A visit identifies the relevant version, 4.0.0, and a contributor list containing Li Peng.
One targeted search for Li Peng as a former Chinese premier checks the identity match, after which Seer submits \emph{Li Peng}.

\textbf{Comparison.}
Base Agent reaches the same ChangeLog, but subsequent queries accumulate contributor names and loosely related keywords rather than checking a specific candidate.
It uses 9 searches and 2 visits without a correct submission.
BoN@4 and AV@4 also fail (39 and 50 real calls), with similar query deterioration in their four trajectories.

\textbf{Interpretation.}
The useful source is available to both Seer and Base Agent; the difference appears in what they do next.
Seer's trajectory preserves the link between the release, its contributors, and the identity clue.
This illustrates the role of next-action guidance after retrieval, rather than an advantage attributable only to finding a better initial source.
The excerpts do not reveal which unexecuted candidates received lower estimator scores.
\end{casebox}

\begin{casebox}{Case 2: Revising a location as constraints accumulate}
\textbf{Setting.} BrowseComp-200; Qwen3.5-397B-A17B.

\textbf{Question.} Identify a scenic hiking location formed during the last ice age, with a body-part-related name, a fall from its viewpoint between 2000 and 2021, and a 2014 search-and-rescue count related to the Grand Canyon's 2022 count by a factor of 84.5.
\textbf{Labeled answer:} Trolltunga.

\textbf{Seer's committed trajectory: 17 searches.}
Early searches consider directions including Angels Landing, the Grand Canyon, and Finger Lakes without establishing a joint match.
A subsequent search connects Trolltunga with the tongue-related name and glacial formation clues.
Further searches check a 2015 fatal fall and the rescue-statistics clue, leading to the answer \emph{Trolltunga}.

\textbf{Comparison.}
Base Agent spends 54 calls exploring partially matching locations and repeatedly searching the rescue-count ratio, then submits \emph{Half Dome}.
BoN@4 and AV@4 use 178 and 124 calls, respectively, without reaching a correct answer.

\textbf{Interpretation.}
Seer's visible trajectory changes direction when another location better fits multiple constraints, whereas Base Agent ultimately settles on a partial match.
This behavior is consistent with comparing alternative next actions before commitment.
It does not imply that Seer retains a persistent tree of location hypotheses: at depth one, each real round starts a new imagined search from the committed history.
The example shows that a large frozen agent can also benefit from guidance over its proposals.
\end{casebox}

\begin{casebox}{Case 3: Refining a lookup rather than rejecting its premise}
\textbf{Setting.} SEAL-0; Qwen3.5-397B-A17B.

\textbf{Question.} According to United Nations data from 2015, what percentage of national parliamentary seats in Northern America were held by women?
\textbf{Labeled answer:} 21.8\%.

\textbf{Seer's committed trajectory: 16 calls (9 searches, 7 visits).}
Initial searches locate UN, IPU, and UNSD material.
Seer moves from broader regional presentations, including \emph{The World's Women 2015}, to a UNdata table containing the Northern America row.
A secondary data presentation supplies a cross-check of the 2015 value, and Seer submits \emph{21.8\%}.

\textbf{Comparison.}
Base Agent encounters the same source ecosystem but questions whether Northern America is reported separately, eventually giving an approximate 22--23\% range after 40 calls.
BoN@4 uses 135 calls and returns approximately 27--28\% with a premise objection; AV@4 uses 104 calls without supplying the exact labeled figure.
Their searches repeatedly return to broader regional groupings rather than resolving the requested row.

\textbf{Interpretation.}
Failing to locate a row in an initial document is not evidence that the requested statistic is unavailable.
Seer's committed actions continue refining the lookup within the relevant data sources instead of substituting an approximate answer or a premise objection.
The contrast concerns action selection under incomplete retrieval, not access to the labeled answer during search.
\end{casebox}

\begin{casebox}{Case 4: Preserving the two-hop structure of a sports lookup}
\textbf{Setting.} XBench-DeepSearch; Qwen3.5-9B.

\textbf{Question.} For the team of the player ranked second in total fouls in the 2023--24 NBA regular season, what was the score difference in its final game that season, including the play-in and playoffs?
\textbf{Labeled answer:} 7 points.

\textbf{Seer's committed trajectory: 8 calls (6 searches, 2 visits).}
Seer searches the regular-season total-foul leaderboard and visits Basketball-Reference to identify Domantas Sabonis and the Sacramento Kings.
It then searches the Kings' final game and visits a game or season page to resolve the relevant score difference.
The submitted answer is \emph{7 points}.

\textbf{Comparison.}
Base Agent initially finds the relevant basketball source, but later queries mix unrelated names and malformed keyword sequences; timeouts also occur.
It ends without an answer after 14 calls.
BoN@4 spends 62 calls and recovers part of the chain without a correct final difference, while AV@4 spends 47 calls and submits $-10$.

\textbf{Interpretation.}
Seer's trajectory keeps the player-to-team and team-to-game links explicit rather than repeatedly reopening inconsistent combinations.
Together with Case 1, this provides a small-agent example of more focused committed queries.
Because the unsuccessful run also contains tool failures, the difference cannot be attributed exclusively to query drift or to any single decision-model component.
\end{casebox}

\subsection{Fewer Real Calls at the Same Answer Quality}

\begin{casebox}{Case 5: Submitting once the comparison is supported}
\textbf{Setting.} GAIA; Qwen3.5-4B.

\textbf{Question.} How many more encoder blocks or layers does BERT-base have than the encoder in \emph{Attention Is All You Need}?
\textbf{Labeled answer:} 6.

\textbf{Seer's committed trajectory: 2 searches, 3 rounds.}
One search establishes that BERT-base has 12 encoder layers, and another establishes that the original Transformer encoder has 6.
Seer then submits $12-6=6$; the submission is a committed round but not a real tool call.

\textbf{Comparison.}
Base Agent also obtains the two layer counts early, but continues checking arXiv, Wikipedia, Hugging Face, and course material before returning the same answer.
It uses 5 searches and 6 visits over 12 rounds, for 11 real calls in total.
Thus, its real-call count is $5.5\times$ Seer's despite identical final correctness.

\textbf{Interpretation.}
The visible difference is the timing of final submission, not the arithmetic.
In Seer, a final-answer action is evaluated by $V_\theta$ with an empty tool response and competes with further tool-using actions (\S\ref{sec:decision-models}).
This supplies a mechanism for selecting submission once the history supports an answer, without a separate stopping threshold.
The example is consistent with this mechanism, but does not establish that the estimator is calibrated to evidence sufficiency.
\end{casebox}

\begin{casebox}{Case 6: Reaching the same evidence without executing four trajectories}
\textbf{Setting.} BrowseComp-200; Qwen3.5-397B-A17B.

\textbf{Question.} A machine used beneath the Colosseum later developed into steam-powered and electric passenger versions. In a 2010s article about accidents involving this machine, who is the sibling of a person catastrophically injured on Thanksgiving?
\textbf{Labeled answer:} Josh / Josh Nelson.

\textbf{Seer's committed trajectory: 2 calls (1 search, 1 visit).}
A search combining the historical-machine and Thanksgiving-injury clues returns a CBS News article about in-home elevator accidents.
A visit identifies the injured person as Jordan Nelson and the sibling as Josh Nelson.
Seer submits the answer \emph{Josh}.

\textbf{Comparison.}
BoN@4 also selects the correct answer through the same essential evidence chain, but its four real trajectories incur 11 searches and 10 visits in aggregate.
Its 21 real calls are $10.5\times$ Seer's count.

\textbf{Interpretation.}
Here the successful evidence chain is short for both methods; the accounting difference includes the real execution of alternative BoN@4 trajectories.
Seer instead uses the simulator and estimator to guide depth-one imagined search, executing only the selected action at each round.
Candidate width $k=32$ is a maximum number of next-action proposals per expansion, not 32 complete simulated research trajectories.
The comparison establishes lower real tool use, not matched local computation or a $10.5\times$ \mbox{latency speedup.}
\end{casebox}

\section{Limitations and Future Work}
\label{app:limitations}

Seer's scope is cost reduction on the tool-call axis, not context management: it reduces how often external services must be called, while the interaction histories it reads still grow with every round.
This section reports the limitations that follow from the design and evaluation choices, states why the current work does not cover each of them, and outlines the corresponding follow-up studies, grouped into five directions: the relation to context compression, model coverage, the question-answering focus, fixed search computation, and deployment cost accounting.

\textbf{Relation to context compression.}
Seer is not a context-management method and does not compete with context compression or summarization.
The two operate on different cost axes: context compression~\citep{ondevicecontext} shrinks what the model reads, whereas Seer reduces how often external services must be called.
Seer keeps full committed histories by design, because the estimator is trained on complete real histories and retaining real responses is what preserves the separation between predicted content and citable evidence.
For the same reason, we did not include a context-compression or summarization baseline in this work: such a method changes the input distribution that the frozen agent sees, so a controlled comparison would require either retraining the agent for compressed histories or verifying that a compression module preserves decision quality on every benchmark, either of which is a study on its own.
Combining Seer with context compression on the same tool interface is a natural direction, as the two would reduce both cost axes at once.

\textbf{Model coverage.}
Our evaluation covers Qwen and Gemma, currently among the most widely used open-source model families, with both decision models initialized from Qwen3.5-9B.
The restriction is deliberate rather than incidental: both decision models are trained once on real trajectories and then shared by every evaluated agent, so extending the study to an independent family such as DeepSeek or GLM means collecting new training trajectories and retraining both decision models on that family, a training-scale effort that would break the controlled setup in which every agent is guided by the same decision models.
The Gemma agents already provide first evidence that the guidance itself transfers across families; whether the shared 9B decision models become a bottleneck for stronger agents, and whether the gains persist when the agent is stronger than the models guiding it, are left to future work.

\textbf{From question answering to research reports.}
We focus on question answering because terminal correctness provides a clear, verifiable training signal and a consistent basis for comparing agents.
Correctness alone does not capture the quality of a full research report, where coverage, citation support, and synthesis also matter.
We did not evaluate report-style tasks in this work because their supervision requires rubrics whose construction is itself an open measurement problem; adopting one would tie both the estimator's training signal and the evaluation to a contested metric.
Extending Seer with rubric-based supervision would preserve the separation between proposal, prediction, and evaluation while broadening what the estimator learns to assess.

\textbf{Search depth and adaptive allocation of computation.}
We use fixed candidate widths and a search depth of one to compare the effect of candidate coverage under a common search procedure; deeper search couples width and depth effects and would confound this comparison.
This controlled setting leaves open how much computation each decision needs, and the estimator's discount factor is likewise fixed at $\gamma=0.95$ throughout, with its sensitivity left uncharacterized.
We did not run an adaptive-computation study because such a policy must commit its budget before observing the outcome, which requires a calibrated uncertainty signal that the current estimator does not provide.
Future work could vary candidate width and simulation effort according to decision uncertainty and the remaining budget, building on adaptive agent scaling~\citep{catts,bats}.
The aim would be to spend more computation where additional candidates are useful and less where they add little, evaluated under the same resource constraints as the fixed settings studied here.

\textbf{Deployment cost accounting on local devices.}
By accounting for model parameters and real tool calls separately, we study the cost of guidance without tying the comparison to a particular device.
These measures do not, however, establish peak memory use, energy consumption, or latency on actual hardware.
We did not take hardware measurements, because they depend on a specific device profile, and single-device numbers would not transfer across the heterogeneous edge devices this work targets.
We instead report a device-independent analytical estimate: each decision proposes $k$ candidates and runs the simulator and estimator once per candidate, so local inference time scales roughly with $k$.
Quantizing the agent and both decision models, sharing representations between prediction and evaluation, and tailoring batching to the device could reduce the added local cost.
Evaluating these changes on a fixed device profile would show whether the savings preserve decision quality and translate into practical deployment benefits.
This design exchanges local computation for performance without agent-side training, and the two decision models share the same 9B base weights with separate LoRA adapters, so the added memory is dominated by the shared base rather than by the adapters.

Progress along these directions could make reusable guidance practical for a wider range of research tasks and devices, accelerating access to research assistance while preserving users' control over their agents' reasoning and decisions.

\end{document}